\documentclass{article}
\PassOptionsToPackage{x11names,dvipsnames,table}{xcolor}

\usepackage{arxiv}
\usepackage{float}
\usepackage[utf8]{inputenc} 
\usepackage[T1]{fontenc}    
\usepackage{hyperref}       
\usepackage{url}            
\usepackage{booktabs}       
\usepackage{amsfonts}       
\usepackage{nicefrac}       
\usepackage{microtype}      
\usepackage{lipsum}
\usepackage{tikz}
\usepackage[numbers]{natbib}
\usepackage{textcomp}
\usepackage{graphicx,color}
\usepackage[x11names,dvipsnames,table]{xcolor}
\usepackage{array}      
\usepackage[utf8]{inputenc}
\usepackage{longtable}
\usepackage{placeins}
\usepackage{afterpage}
\usepackage{pgfplots}
\pgfplotsset{compat=1.18}
\usepackage[english]{babel}
\usepackage{amsmath}
\usetikzlibrary{shapes.geometric, arrows.meta,chains, positioning, fit}
\usepackage{caption}
\usepackage{siunitx}
\graphicspath{ {./images/} }

\title{Continuous Saudi Sign Language Recognition:  A Vision Transformer Approach}

\author{
 Soukeina Elhassen \\
  Department of Computer Science, Faculty of Computing and Information Technology\\
  King Abdulaziz University\\
  Jeddah 21589, Saudi Arabia\\
  \texttt{selhassen@stu.kau.edu.sa} \\
   \And
 Lama Al Khuzayem \\
  Department of Computer Science, Faculty of Computing and Information Technology\\
  King Abdulaziz University\\
  Jeddah 21589, Saudi Arabia\\
  \And
 Areej Alhothali \\
  Department of Computer Science, Faculty of Computing and Information Technology\\
  King Abdulaziz University\\
  Jeddah 21589, Saudi Arabia\\
    \AND
  Ohoud Alzamzami \\
  Department of Computer Science, Faculty of Computing and Information Technology\\
  King Abdulaziz University\\
  Jeddah 21589, Saudi Arabia\\
  \And
  Nahed Alowaidi\\
  Department of Computer Science, Faculty of Computing and Information Technology\\
  King Abdulaziz University\\
  Jeddah 21589, Saudi Arabia\\
}

\begin{document}
\maketitle
\begin{abstract}
Sign language (SL) is an essential communication form for hearing-impaired and deaf people, enabling engagement within the broader society. Despite its significance, limited public awareness of SL often leads to inequitable access to educational and professional opportunities, thereby contributing to social exclusion, particularly in Saudi Arabia, where over 84,000 individuals depend on Saudi Sign Language (SSL) as their primary form of communication. Although certain technological approaches have helped to improve communication for individuals with hearing impairments, there continues to be an urgent requirement for more precise and dependable translation techniques, especially for Arabic sign language variants like SSL. Most state-of-the-art solutions have primarily focused on non-Arabic sign languages, resulting in a considerable absence of resources dedicated to Arabic sign language, specifically SSL. The complexity of the Arabic language and the prevalence of isolated sign language datasets that concentrate on individual words instead of continuous speech contribute to this issue. To address this gap, our research represents an important step in developing SSL resources. To address this, we introduce the first continuous Saudi Sign Language dataset called KAU-CSSL, focusing on complete sentences to facilitate further research and enable sophisticated recognition systems for SSL recognition and translation. Additionally, we propose a transformer-based model, utilizing a pretrained ResNet-18 for spatial feature extraction and a Transformer Encoder with Bidirectional LSTM for temporal dependencies, achieving 99.02\% accuracy at signer dependent mode and 77.71\% accuracy at signer independent mode. This development leads the way to not only improving communication tools for the SSL community but also making a substantial contribution to the wider field of sign language.
\end{abstract}


\section{Introduction}

By 2050, the number of people who will suffer from either hearing loss or hard of hearing is $2.5$ billion, according to the World Health Organization~\cite{who2024deafness}, a global challenge that underscores the need for effective communication solutions ~\cite{Wadhawan2020}. In Saudi Arabia, the General Population and Housing Census~\cite{apd_statistics} estimated the number of hearing-impaired people in $2022$ at $84,025$. This indicates that more than $84,000$ people use Saudi Sign Language (SSL) as an essential method of communication in the Saudi Arabian hard-of-hearing community ~\cite{Luqman2021}. However, beyond the deaf and hard-of-hearing community, the understanding of SSL is limited, creating significant barriers for individuals seeking essential services such as education and healthcare, thereby contributing to their social isolation ~\cite{Algethami2025}.

Sign Language recognition (SLR) systems that translate sign language (SL) gestures into text or speech are essential for bridging this communication gap ~\cite{Rastgoo2020}. These systems facilitate real-time interaction between the hearing-impaired community and the broader society, enhancing accessibility to essential services. Robust SLR systems need to be trained on high-quality datasets that cover real-world communication scenarios~\cite{Luqman2022}. Without accurate and robust recognition systems, miscommunication and misunderstanding can occur, which might lead to further isolation from the broader community. Thus, the development of the SSL dataset represents a substantial advancement in the development of accurate and robust translation models.

The task of SLR is inherently difficult due to variations in hand motion, pace, shape, and position ~\cite{Balaha2023}. Further challenges arise from differences in signing styles, such as left-handed or right-handed signers. In addition, the signer's hands may obscure facial expressions, which introduces additional complexity. Moreover, the linguistic lexicon of SL increases the number of classes to be covered, which might result in several classes equal to the complete dictionary; in addition, a single word may have multiple different gestures. For instance, the word 'doctor' can be signed in two different ways, making it challenging to cover the entire lexicon. These challenges are particularly acute in continuous SSL, where transitions between signs and contextual nuances must be captured ~\cite{Alkhalifa2025}.


The SLR task can be formulated as an image-based, isolated-video-based, and continuous-video-based method. Image-based methods typically focus on recognizing static hand shapes and configurations using cropped images of a signer’s hands performing a specific sign \cite{al2021deep}. These methods are often used to identify individual signs or letters, but often lack the dynamic context necessary for understanding full SL expressions. 

Isolated-video-based recognition approaches, on the other hand, focus on recognizing a single sign from video clips, where signers perform predefined signs in a controlled environment. This approach is more advantageous than the image-based approach for its ability to capture temporal information, such as hand movement and shape changes, providing a more dynamic representation than image-based methods. However, this approach operates within an individual isolated sign and cannot capture the smooth transition between signs that occur in natural continuous signing~\cite{al2021deep}.

The continuous-based method involves recognizing sequences of signs within longer videos, where signers perform multiple signs or full sentences. Continuous-based datasets more accurately reflect the difficulties of SL and present the real-world challenges of SL, including transitions between signs, variations in pace, and facial expressions, which allows for the development of more robust models ~\cite{Batnasan2022}. Despite this, most state-of-the-art SLR research targets isolated datasets due to the scarcity of continuous ones~\cite{Elbadawy2024}.

Globally, SLR task has an active research area with researchers developing models to recognize SLs associated with various spoken languages, such as American Sign Language (ASL), British Sign Language (BSL), and Chinese Sign Language (CSL)~\cite{Ahmed2025}. Despite the progress made in these areas, the research on SSL has been relatively limited compared to the other languages. Most of existing studies in Arabic focus on the Arabic Sign Language (ArSL), which is a generalized sign language used in various Arab-speaking countries \cite{adaloglou2021comprehensive}.

Significantly less attention has been devoted to SSL, despite its extensive use within the hearing-impaired community in Saudi Arabia. In addition, the research effort in SSL focuses only on image-based and isolated video approaches, and no previous studies have tackled the problem of SLR in the context of continuous SSL \cite{alsulaiman2023facilitating}.

To address this research gap, we developed the first continuous SSL dataset that called KAU-CSSL. The dataset consists of videos showing individuals performing SSL of medical-related sentences. This dataset is intended to pave the way for groundbreaking investigations into continuous SSL recognition and translation. We also designed a transformer-based model, leveraging a pretrained ResNet-18 for spatial feature extraction and a Transformer Encoder with Bidirectional LSTM for temporal dependencies, trained on KAU-CSSL to achieve 99.02\% accuracy in signer dependent mode and 77.71\% in signer independent mode where it must generalize to unseen signers, highlighting its potential for real-world applications despite greater challenges~\cite{BaniBaker2023}. This contribution represents a critical milestone in the progression of SL technology, with the potential to fundamentally transform the accessibility of communication for individuals with hearing impairments not only in Saudi Arabia but also globally.

\textbf{Key Contributions of This Work:}
\begin{itemize}
    \item Development of the KAU-CSSL dataset, the first benchmark dataset for continuous Saudi Sign Language (SSL) recognition, with 5,810 videos across 85 medical-related classes, addressing a critical gap in Arabic SL resources. 
    \item Development of "KAU-SignTransformer" a transformer-based model achieving 99.02\% accuracy in signer-dependent mode, it attains 77.71\% accuracy in signer-independent mode , demonstrating  the effectiveness of transformer network for continuous sign language recognition tasks.
    \item An ablation study to evaluate the impact of model components (e.g., pretrained ResNet-18, Bidirectional LSTM) on performance.
\end{itemize}

\section{Related Work}
Sign language recognition (SLR) has emerged as a significant area of study for enhancing accessibility to hard-of-hearing and Deaf individuals, particularly by methods of real-time SLR systems converting natural, sentence-level signing into text and spoken language. While online SLR for languages like American Sign Language (ASL) and German Sign Language (DGS) has progressed, there is minimal resources for Arabic Sign Language (ArSL), notably Saudi Sign Language (SSL). This shortage is particularly notable for continuous SSL datasets, which are vital in real-world situations like healthcare communication in Saudi Arabia. The lack of such resourses slows the development of robust models that can handle the complexities of continuous signing like movement epenthesis and signer variation. This section describes existing continuous SL datasets and SL recognition approaches, their merits, and their limitations.

\paragraph{Continuous Datasets}
Continuous sign language recognition (SLR) relies on comprehensive datasets that capture natural, sentence-level signing across various sign languages (SLs), such as American Sign Language (ASL), British Sign Language (BSL), and Chinese Sign Language (CSL). These datasets are essential for developing models that can interpret the fluid, context-dependent nature of real-world signing, which often includes co-articulation and non-manual features like facial expressions. Notable datasets include RWTH-PHOENIX-Weather 2014~\cite{forster2012rwth}, which focuses on German Sign Language (DGS) with weather-related content, collected under controlled conditions to ensure consistency across 9 signers and 8,257 video samples, addressing the challenge of temporal segmentation in continuous sequences. SIGNUM~\cite{agris2010signum}, another pivotal DGS dataset, comprises 14,000 video samples from 25 signers, designed to support research into dynamic gesture recognition by incorporating varied signing speeds and styles, though it faces limitations due to its constrained vocabulary of 450 signs. How2Sign~\cite{duarte2021how2sign}, a large-scale ASL dataset with 35,000 videos from 11 signers, emphasizes instructional contexts and includes multi-modal data (e.g., video, gloss annotations), making it a valuable resource for training models on complex, multi-sentence signing, despite challenges in handling signer variability and background noise. These datasets have been instrumental in advancing continuous SLR models by providing diverse and annotated video samples for training and evaluation, enabling researchers to tackle issues like movement epenthesis and contextual dependencies.

In Arabic Sign Language (ArSL), the landscape is less developed, with most available datasets, such as the KING SAUD UNIVERSITY SAUDI SIGN LANGUAGE (KSU-SSL) dataset~\cite{al2020hand} for SSL, limited to isolated signs, focusing on individual gestures rather than continuous sequences. This isolation restricts their applicability to real-time, sentence-level communication, a critical need for accessibility in Arabic-speaking regions. However, ArabSign~\cite{luqmanArabsign2023}, an ArSL dataset with 50 continuous sentences and 9,335 video samples from 6 signers, represents a step toward addressing this gap by offering a modest foundation for continuous recognition, though its small class size and limited signer diversity pose challenges for robust generalization. Despite this progress, continuous SSL datasets remained unavailable prior to our work, highlighting a significant barrier to developing tailored SLR systems for Saudi Arabia’s Deaf and hard-of-hearing community, where regional sign variations and medical communication needs are particularly acute. Our study introduces the KAU-CSSL dataset, a novel continuous SSL dataset designed for medical contexts, addressing this deficiency by providing 5,810 video samples across 85 classes, thereby enabling advanced research into continuous SSL recognition. Table~\ref{tab:Cont-datasets} summarizes publicly available continuous SL datasets, highlighting the diversity and scope of existing resources and underscoring the unique contribution of KAU-CSSL.
\begin{table*}[h]
\centering
\caption{Summary of Publicly Available Continuous Sign Language Datasets}
\renewcommand{\arraystretch}{1.2}
\small
\resizebox{\textwidth}{!}{%
\begin{tabular}{|c|c|c|c|c|c|}
\hline
\textbf{Dataset} & \textbf{Year} & \textbf{Language} & \textbf{Classes} & \textbf{\#Signers} & \textbf{\#Samples} \\ \hline
SIGNUM               & 2007 & German Sign Language (DGS)                              & 450   & 25  & 14000  \\ \hline
RWTH-BOSTON-104      & 2008 & American Sign Language (ASL)                            & 168   & 3   & 201    \\ \hline
RWTH-BOSTON-400      & 2008 & American Sign Language (ASL)                            & 406   & 4   & 843    \\ \hline
ATIS                 & 2008 & BSL, DGS, ISL, SASL                                     & 400   & 6   & 595    \\ \hline
Corpus-NGT           & 2008 & Sign Language of the Netherlands (NGT)                  & 3900  & 92  & 116    \\ \hline
RWTH-PHOENIX-weather & 2012 & German Sign Language (DGS)                              & 911   & 7   & 1980   \\ \hline
RWTH-PHOENIX-weather & 2014 & German Sign Language (DGS)                              & 1558  & 9   & 8257   \\ \hline
HOW2SIGN             & 2021 & American Sign Language (ASL)                            & 16000 & 11  & --     \\ \hline
LSA-T                & 2021 & Argentine Sign Language (LSA)                           & --    & 103 & 14880  \\ \hline
LSFB-CONT            & 2021 & French Belgian Sign Language (LSFB)                     & 6883  & 100 & +85000 \\ \hline
NCSLGR               & 2022 & American Sign Language (ASL)                            & 1800  & 4   & --     \\ \hline
ArabSign             & 2023 & Arabic Sign Language (ArSL)                             & 50    & 6   & 9335   \\ \hline
\textbf{KAU-CSSL}    & \textbf{2025} & \textbf{Saudi Sign Language (SSL)}                     & \textbf{85}   & \textbf{24} & \textbf{5810} \\ \hline
\end{tabular}%
}
\label{tab:Cont-datasets}
\end{table*}

\paragraph{Saudi Sign language Recognition Approaches}
\label{subsec:Saudi-LR-title}

There are limited number of research published in the field of SSL recognition. To the best of our knowledge these are the only papers in SSL. Al-Obodi et al.~\cite{alsaudi} used CNN to handle the SSL recognition problem, in which they used their own dataset which consists of RGB images of the signs that contain only one gesture. The researchers achieved an accuracy of 99.47\%.
Al-Hammadi et al.~\cite{al2019hand} proposed a model using 3DCNN on RGB videos. The researchers evaluated the model using the KSU-SSL dataset along with other datasets. The proposed model employs transfer learning to overcome the lack of annotation, where the researchers used a 3DCNN model that had been trained on human action recognition task. The authors have employed two models, a single 3DCNN where the feature extraction phase is done on the entire video samples, and a parallel 3DCNN where the temporal dependency is enhanced by extracting features from different regions from the video samples. They used different features fusion techniques which include, MLP fusion, LSTM fusion, and AUTOENCODER in the parallel 3DCNN. The results show that the parallel 3DCNN outperforms the single 3DCNN where the single 3DCNN achieved 72.32\% while the parallel 3DCNN model achieved 82.19\% with the LSTM and 84.38\% with the MLP fusion architecture. 

Al-Hammadi et al.~\cite{al2020deep} proposed a model that addresses some the limitations in their previous work~\cite{al2019hand}. They used open pose framework to detect 2D key points, then the model learns spatiotemporal features using a model that was on a human action recognition dataset. MLP and autoencoder have been used as feature fusion techniques. The proposed model at the signer-independent mode achieved 84.89\% with autoencoder and 87.69\% with MLP.  

Abdul et al.~\cite{abdul2021intelligent} proposed a model that uses a CNN Inception model with an attention mechanism and BiLSTM. Inception-v3 was used to extract the spatial features, while Bi-LSTM was utilized to extract the temporal features from the video sequences. Three dynamic datasets were used to evaluate the model, which are the KSU-ArSL dataset\cite{al2020hand}, the Jester dataset\cite{materzynska2019jester}, the NVIDIA gesture database and achieved accuracies of 84.2\%, 95.8\%, and 86.6\%, respectively.

Al-Hammadi et al.~\cite{al2022spatial} proposed a model that uses a convolutional graph neural network (GCN) along with a spatial attention mechanism. The researchers used the MediaPipe library to construct the graph and used multi-head self-attention to improve context representation at every frame. The model was evaluated on different datasets, which are the KSU-SSL dataset, AUTSL, ASLLVD-20, ASLLVD, LSA-64, and Jester Dataset. The model achieve 90.22\%, 91.6\%, 30.47\%, 68.75\%, 94.22\% and 80.3\% accuracy respectively.

An isolated SSL dataset has been constructed by Al-Mohimeed et al.~\cite{al2022dynamic}, the dataset consists of thirty-five words at health queries and common words such as "I am, Feeling, Hospital". The researchers proposed a model that uses convolutional long short-term memory (convLSTM) to perform recognition tasks, the model achieved accuracy of 70\%.

Al Khuzayem et al.~\cite{alkhuzayem2024} recently introduced "Efhamni," a mobile application for isolated SSL recognition, developed by the Department of Computer Science at King Abdulaziz University. It uses a CNN-BiLSTM model to translate isolated SSL gestures into Arabic text and voice, serving as a bridge for the deaf and hard-of-hearing community in Saudi Arabia to communicate.  Efhamni was evaluated on a large dataset, achieving superior performance compared to several state-of-the-art approaches, with comparable results to current approaches. Their system utilizes pose estimation and MobileNet for feature extraction, and usability testing confirmed its applicability in the context of deaf and hard-of-hearing users. Efhamni focus on isolated signs, which limits its application to sentence-level recognition. The authors plan to enhance model accuracy and incorporate additional features in the future, emphasizing the value of datasets like KAU-CSSL.

Most of the proposed models in the field are either image-based or isolated video-based SL recognition approaches. Non-Arabic research has addressed the problem of continuous SL recognition in a limited number of studies and has not yet attained optimal results. Thus, there is still much room for performance improvement in continuous SL recognition. Much of the potential future direction in this field include taking into account the attention mechanism, utilizing multiple input modalities to gain the most from multi-channel information, learning structured spatial-temporal patterns , and applying the prior knowledge of SL.
On the other hand, to the best of our knowledge, Saudi research did not address continuous SL recognition, despite its importance in real-time applications.
 
\section{KAU-CSSL Dataset}

The collected dataset, known as "KAU-CSSL" dataset. The dataset consists of videos that represents SL sentences. The KAU-CSSL dataset offers distinctive contributions to the field of sign language recognition while exhibiting significant similarities to popular datasets and responding to the requirements of continuous signing and organized recording. \textbf{Structure and Methodology} KAU-CSSL uses a controlled and standardized recording procedure, much to the How2Sign dataset~\cite{duarte2021how2sign}. To guarantee consistency in gestures and signing sequences, participants watch reference videos that have already been recorded. By reducing variability, this process improves the quality of the dataset and makes it more appropriate for machine learning applications. Furthermore, by incorporating different signers and numerous recordings of every sentence, KAU-CSSL takes a comprehensive approach to diversity, which enhances the dataset's resilience and ability to handle differences in signing techniques. \textbf{Diversity and Generalization} The dataset features multiple recordings of every statement and signers from a variety of backgrounds to further improve its usefulness and generalizability. These characteristics enable KAU-CSSL to be stable and comparable to large-scale datasets such as RWTH-PHOENIX-Weather 2014T ~\cite{forster2012rwth} and How2Sign, facilitating a wide range of research applications and allowing for efficient generalization across various signature methods. \textbf{Domain-Specific Focus:} While How2Sign focuses on instructional content and datasets like RWTH-PHOENIX-Weather 2014 focus on weather-related terminology, KAU-CSSL is tailored for the healthcare sector, filling an important gap in domain-specific sign language resources. Because of this specialty, it is especially useful for creating sign language recognition systems in medical contexts where precise medical terminology communication is crucial.

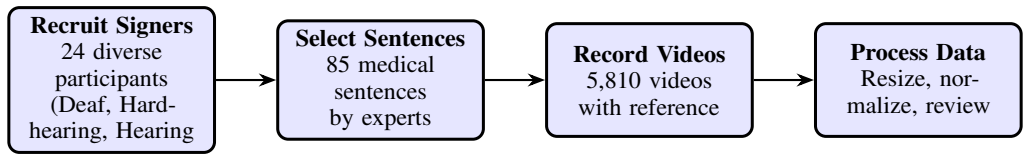
\begin{figure}[ht] \centering \begin{tikzpicture}[ scale=0.8, node distance=1.5cm and 0.8cm, start chain=going right, every node/.style={ rectangle, rounded corners, draw, text width=2.5cm, minimum height=1.5cm, align=center, font=\footnotesize, fill=blue!10, line width=0.4mm }, arrow/.style={ -Stealth, thick } ]

\node [on chain] (recruit) {\textbf{Recruit Signers} \ 24 diverse participants (Deaf, Hard-hearing, Hearing}; \node [on chain] (select) {\textbf{Select Sentences} \ 85 medical sentences by experts}; \node [on chain] (record) {\textbf{Record Videos} \ 5,810 videos with reference}; \node [on chain] (process) {\textbf{Process Data} \ Resize, normalize, review};

\draw[arrow] (recruit) -- (select); \draw[arrow] (select) -- (record); \draw[arrow] (record) -- (process);

\end{tikzpicture} \caption{Pipeline of KAU-CSSL dataset creation: Recruit Signers, Select Sentences, Record Videos, and Process Data.} \label{fig:kaucssl_pipeline} \end{figure}

The creation process of the KAU-CSSL dataset, consist of 4 phases, Recruit Signers, Select Sentences, Record Videos, and Process Data. Figure~\ref{fig:kaucssl_pipeline} shows a general Pipeline of KAU-CSSL dataset creation.

\paragraph{Phase 1: Recruit Signers}

\begin{figure}[H]
\centering
\begin{tikzpicture}
\definecolor{purple}{HTML}{A02B93}
\definecolor{darkblue}{HTML}{0E2841}
\definecolor{teal}{HTML}{116287}
\begin{axis}[
    title={No. of Videos per Class},
    title style={font=\small, yshift=-1ex},
    ybar=0.15cm,
    width=8cm,
    height=5cm,
    bar width=0.5cm,
    symbolic x coords={Maximum, Minimum, Average},
    xtick={Maximum, Minimum, Average},
    xticklabels={Maximum, Minimum, Average},
    xticklabel style={font=\small, anchor=north, inner sep=1pt},
    enlarge x limits={0.4},
    ymin=0, ymax=90,
    ylabel={No. of Videos},
    ylabel style={font=\small},
    xlabel style={font=\small},
    yticklabel style={font=\small},
    nodes near coords,
    nodes near coords style={font=\small},
    axis line style={-},
    grid style={dashed, gray!30},
]
\addplot[fill=purple] coordinates {(Maximum, 74)};
\addplot[fill=darkblue] coordinates {(Minimum, 63)};
\addplot[fill=teal] coordinates {(Average, 68)};
\end{axis}
\end{tikzpicture}
\caption{Number of Videos per Class in KAU-CSSL: Maximum (74), Minimum (63), Average (68).\label{fig:no_vid}}
\end{figure}
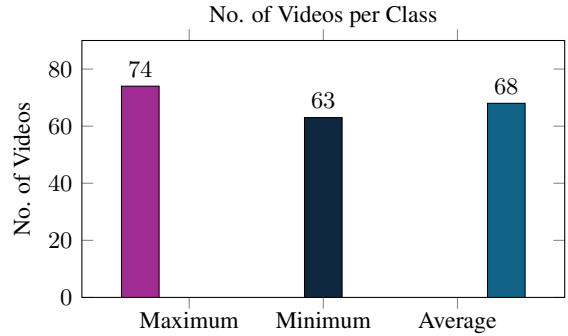

The total number of participants was $24$ signers. The signers were a diverse group of hearing-impaired people, expert translators, and non-experts, with various hearing ability, race, gender and age. The KAU-CSSL Dataset contains $85$ distinct sentences, which comprise a lexicon of roughly $5,000$ words. Consequently, the dataset consists of a total of $5,810$ videos, each of the $24$ signers executed each sentence three times. The maximum, minimum, and average number of videos in a class are 74, 63, and 68, respectively. Figure~\ref{fig:no_vid} illustrates the number of videos per class.
 
\par
\begin{figure}[]
\centering
    \includegraphics[width=8 cm]{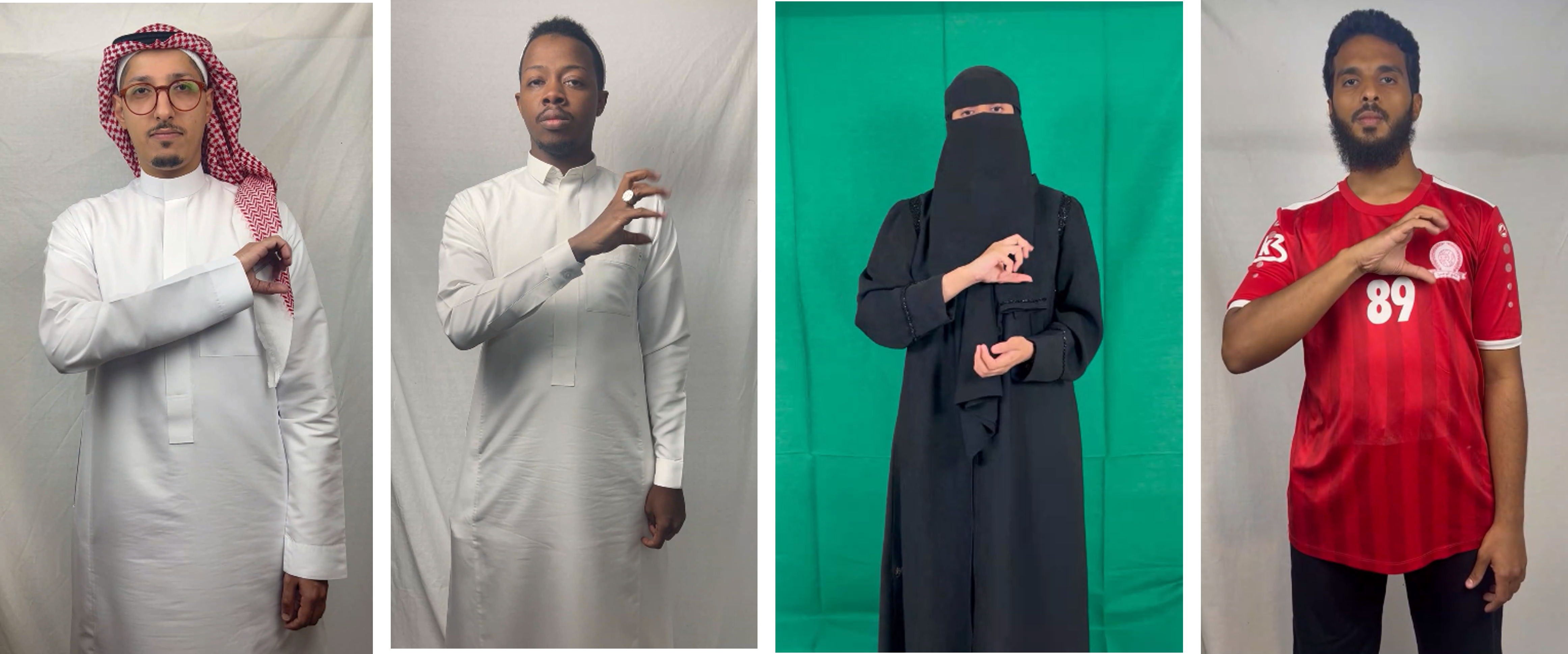}
    \caption{Sample of the dataset}
    \label{fig:dataset_diversity}
\end{figure}

\textbf{Diversity in Participants} The compilation of the dataset prioritized inclusivity and diversity, with the aim of capturing a broad range of SL expressions reflective of real-world medical communication scenarios. To accomplish this, the dataset included a diverse group of participants with a variety of attributes. Figure \ref{fig:dataset_diversity} shows a sample of dataset diversity.
\par
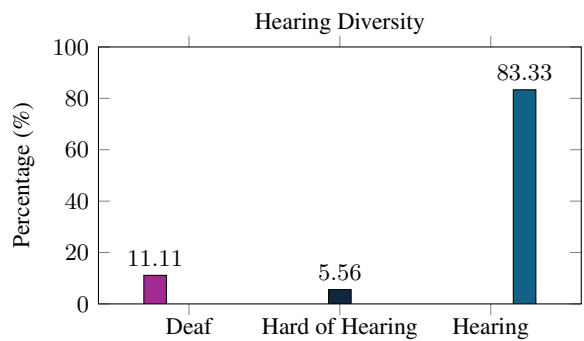
\begin{figure}[H]
\centering
\begin{tikzpicture}
\definecolor{purple}{HTML}{A02B93}
\definecolor{darkblue}{HTML}{0E2841}
\definecolor{teal}{HTML}{116287}
\begin{axis}[
    title={Hearing Diversity},
    title style={font=\small, yshift=-1ex},
    ybar=0.15cm,
    width=8cm,
    height=5cm,
    bar width=0.3cm,
    symbolic x coords={Deaf, Hard of Hearing, Hearing},
    xtick={Deaf, Hard of Hearing, Hearing},
    xticklabels={Deaf, Hard of Hearing, Hearing},
    xticklabel style={font=\small, anchor=north, inner sep=1pt},
    enlarge x limits={0.3},
    ymin=0, ymax=100,
    ylabel={Percentage (\%)},
    ylabel style={font=\small},
    xlabel style={font=\small},
    yticklabel style={font=\small},
    nodes near coords,
    nodes near coords style={font=\small},
    axis line style={-},
    grid style={dashed, gray!30},
]
\addplot[fill=purple] coordinates {(Deaf, 11.11)};
\addplot[fill=darkblue] coordinates {(Hard of Hearing, 5.56)};
\addplot[fill=teal] coordinates {(Hearing, 83.33)};
\end{axis}
\end{tikzpicture}
\caption{Hearing Diversity in KAU-CSSL: Deaf (11.11\%), Hard of Hearing (5.56\%), Hearing (83.33\%).\label{fig:Hearing_Diversity}}
\end{figure}

\textbf{Diversity in Hearing Abilities} The dataset contains a diverse group of participants, including deaf, hard-of-hearing, and hearing individuals. This diversity was crucial for capturing an exhaustive range of SL expressions and ensuring that the dataset is representative of actual communication scenarios. Figure~\ref{fig:Hearing_Diversity} illustrates the diversity of hearing ability among the participants.
 
\par
\textbf{Diversity in Skin Tone} To build a robust SLR, it is important to consider cultural and ethnic diversity in SL datasets. Therefore, we incorporated individuals of various complexion tones in the data collection stage. By including participants with various skin tones and race, the dataset aimed to capture the diversity and inclusiveness inherent in SL communication across diverse backgrounds.
\par
\begin{figure}[H]
\centering
\begin{tikzpicture}
\definecolor{purple}{HTML}{A02B93}
\definecolor{darkblue}{HTML}{0E2841}
\begin{axis}[
    title={Gender Diversity},
    title style={font=\small, yshift=-1ex},
    ybar,
    width=10cm,
    height=5cm,
    bar width=0.9cm,
    symbolic x coords={Male, Female},
    xtick={Male, Female},
    enlarge x limits={0.9},
    ymin=0, ymax=100,
    ylabel={Percentage (\%)},
    ylabel style={font=\small},
    xlabel style={font=\small},
    xticklabel style={font=\small},
    yticklabel style={font=\small},
    nodes near coords,
    nodes near coords style={font=\small},
    axis line style={-},
    grid style={dashed, gray!30},
]
\addplot[fill=purple] coordinates {(Male, 65)};
\addplot[fill=darkblue] coordinates {(Female, 35)};
\end{axis}
\end{tikzpicture}
\caption{Gender Diversity in KAU-CSSL Dataset: Male (65\%) and Female (35\%) Signers.\label{fig:gender_diversity}}
\end{figure}
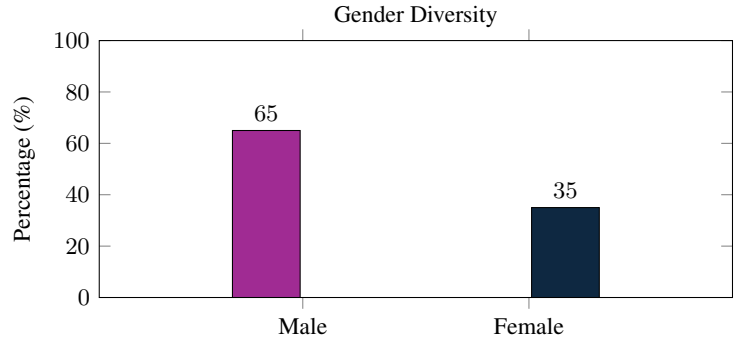

\textbf{Gender Diversity} Significant emphasis was placed on gender equality, and both men and women were invited to participate. This method ensured that the dataset is representative of the various demographics encountered in healthcare settings and reflected the contribution of gender identities in medical communication. Figure~\ref{fig:gender_diversity} illustrates gender diversity.
 
\textbf{Attire Diversity} In pursuit of comprehensive representation and authenticity, the dataset was gathered with a focus on attire diversity, where we focused on the participation of women who wore the "Niqab" that is a face-covering garment. A part of women participant only wore hear cover without Niqab, and a part of men wore the traditional "Thob". This decision is based on the realization that traditional attire is an integral part of the identities and daily lives of people in Saudi Arabia, and its incorporation in the dataset has significant implications for its practical application. However, this decision presents challenges that must be overcome.
Figures~\ref{fig:attire_diversity_female} and ~\ref{fig:attire_diversity_male} illustrate diversity of attire among male and female participants.

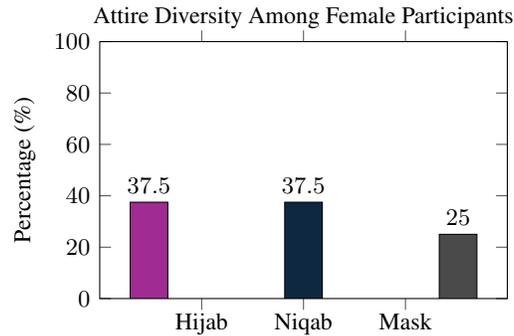
\begin{figure}[H]
\centering
\begin{tikzpicture}
\definecolor{purple}{HTML}{A02B93}
\definecolor{darkblue}{HTML}{0E2841}
\definecolor{gray}{HTML}{4A4A4A}
\begin{axis}[
    title={Attire Diversity Among Female Participants},
    title style={font=\small, yshift=-1ex},
    ybar=0.2cm,
    width=7cm,
    height=5cm,
    bar width=0.5cm,
    symbolic x coords={Fully Visible Face (Hijab), Eyes Only Visible (Face Niqab), Face Mask},
    xtick={Fully Visible Face (Hijab), Eyes Only Visible (Face Niqab), Face Mask},
    xticklabels={Hijab, Niqab, Mask},
    xticklabel style={font=\small, anchor=north, inner sep=1pt},
    enlarge x limits={0.5},
    ymin=0, ymax=100,
    ylabel={Percentage (\%)},
    ylabel style={font=\small},
    xlabel style={font=\small},
    yticklabel style={font=\small},
    nodes near coords,
    nodes near coords style={font=\small},
    axis line style={-},
    grid style={dashed, gray!30},
]
\addplot[fill=purple] coordinates {(Fully Visible Face (Hijab), 37.50)};
\addplot[fill=darkblue] coordinates {(Eyes Only Visible (Face Niqab), 37.50)};
\addplot[fill=gray] coordinates {(Face Mask, 25.00)};
\end{axis}
\end{tikzpicture}
\caption{Attire Diversity Among Female Participants in KAU-CSSL: Fully Visible Face (Hijab) (37.50\%), Eyes Only Visible (Face Niqab) (37.50\%), Face Mask (25.00\%).\label{fig:attire_diversity_female}}
\end{figure}
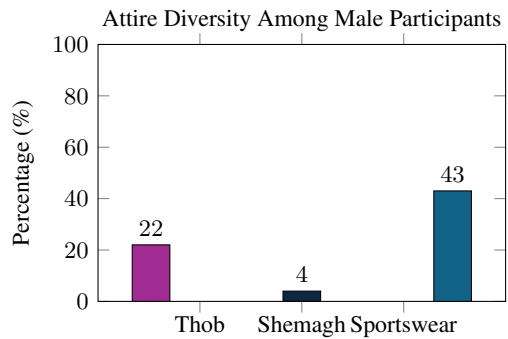
\begin{figure}[H]
\centering
\begin{tikzpicture}
\definecolor{purple}{HTML}{A02B93}
\definecolor{darkblue}{HTML}{0E2841}
\definecolor{teal}{HTML}{116287}
\begin{axis}[
    title={Attire Diversity Among Male Participants},
    title style={font=\small, yshift=-1ex},
    ybar=0.15cm,
    width=7cm,
    height=5cm,
    bar width=0.5cm,
    symbolic x coords={Traditional Thob, Head Cover (Shemagh/Shawl), Sportswear},
    xtick={Traditional Thob, Head Cover (Shemagh/Shawl), Sportswear},
    xticklabels={Thob, Shemagh, Sportswear},
    xticklabel style={font=\small, anchor=north, inner sep=1pt},
    enlarge x limits={0.5},
    ymin=0, ymax=100,
    ylabel={Percentage (\%)},
    ylabel style={font=\small},
    xlabel style={font=\small},
    yticklabel style={font=\small},
    nodes near coords,
    nodes near coords style={font=\small},
    axis line style={-},
    grid style={dashed, gray!30},
]
\addplot[fill=purple] coordinates {(Traditional Thob, 22)};
\addplot[fill=darkblue] coordinates {(Head Cover (Shemagh/Shawl), 4)};
\addplot[fill=teal] coordinates {(Sportswear, 43.00)};
\end{axis}
\end{tikzpicture}
\caption{Attire Diversity Among Male Participants in KAU-CSSL: Traditional Thob (22\%), Head Cover (Shemagh/Shawl) (4\%), Sportswear (43.00\%).\label{fig:attire_diversity_male}}
\end{figure}

KAU-CSSL Dataset is unique in that it is the first continuous video dataset to include SSL. This pioneering effort represents a significant technological and research advancement in the field of SL.

\paragraph{Phase 2: Select Sentences}
KAU-CSSL dataset consists of recordings SL sentences intended for medical communication in a variety of scenarios. The collected dataset is based on the Saudi Dictionary, which was produced and published in 2018 by the Saudi Association for Hearing Disability~\cite{DictionarySaudiSignLanguage}. The dictionary contains hundreds of signs in several categories, including religious, medical, and social, among others.

The initial phase of the creation focused on gathering a wide range of sentences that covered various scenarios in a hospital setting, including medical emergencies, interactions with hospital staff, administrative procedures, and outpatient clinic visits. A SL translator that specializing in medical terminology was then engaged to refine and simplify these sentences for better comprehension and usability by deaf patients. To refine the selected sentences, a series of stringent criteria were used in the refinement processes:
 \textbf{Conciseness} Sentences must be short, mostly we used three to seven words, but still made sure they represent the meaning we aim to deliver. \textbf{Simplicity} No slang or complex structures, where the sentences should be easy to follow and the language should be simple.
 \textbf{Significance} The sentences must focus on vocabulary and expressions about general hospital scenarios.
 \textbf{Clarity} The selected words must be unambiguous and without the possibility of multiple interpretations.
\textbf{Consistency with Deaf Communication Patterns} All sentences must comply with typical deaf communication styles that is subject-verb-object word order and no filler words is used.
 \textbf{Alignment with Saudi Sign Language Vocabulary} Selected words and sentences will be aligned with the SSL vocabulary, focusing on signs that are documented to ensure increased recognition of the signs. This approach results in both consistency and familiarity for the Saudi hearing impaired, as we are avoiding terms without locally recognized or locally accepted.

\begin{figure}[]
\centering
    \includegraphics[width=8 cm]{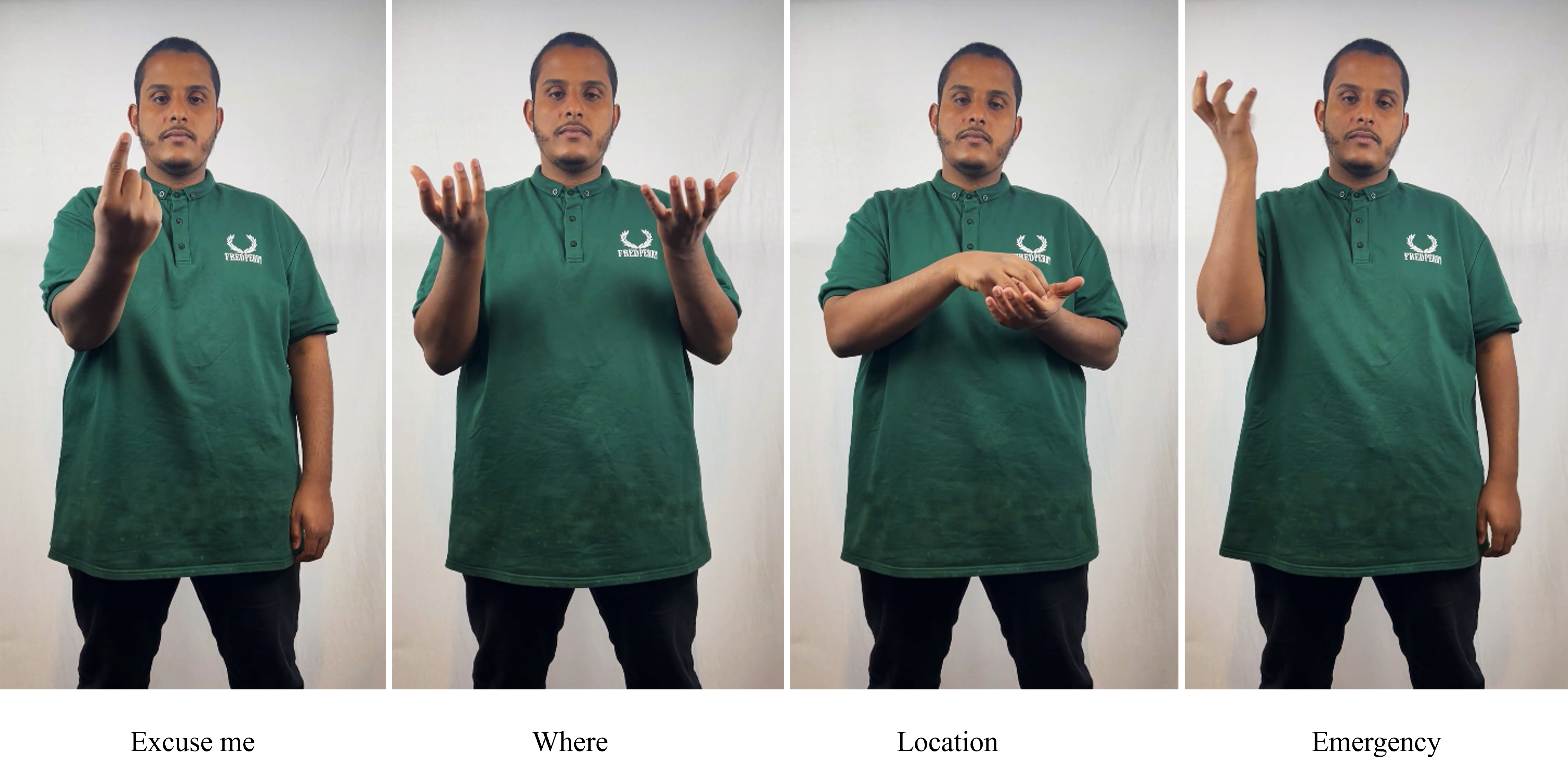}
    \caption{Sample Sentence}
    \label{fig:where_er}
\end{figure}

The primary goal was to facilitate effective communication between hospital-based healthcare professionals and deaf patients. This collaboration guaranteed the practical applicability and appropriateness of the sentences. The final sentence consists of three to five words, such as "Where is the pharmacy?", "I need a doctor", "I want general Practitioner", "I have breathing difficulty", "Where is the location of radiology department", "I need urgent assistance", and other sentences that will increase the quality of the services offered.
Figure~\ref{fig:where_er} illustrates a sentence from the dataset.

\begin{figure}[]
\centering
    \includegraphics[width=8 cm]{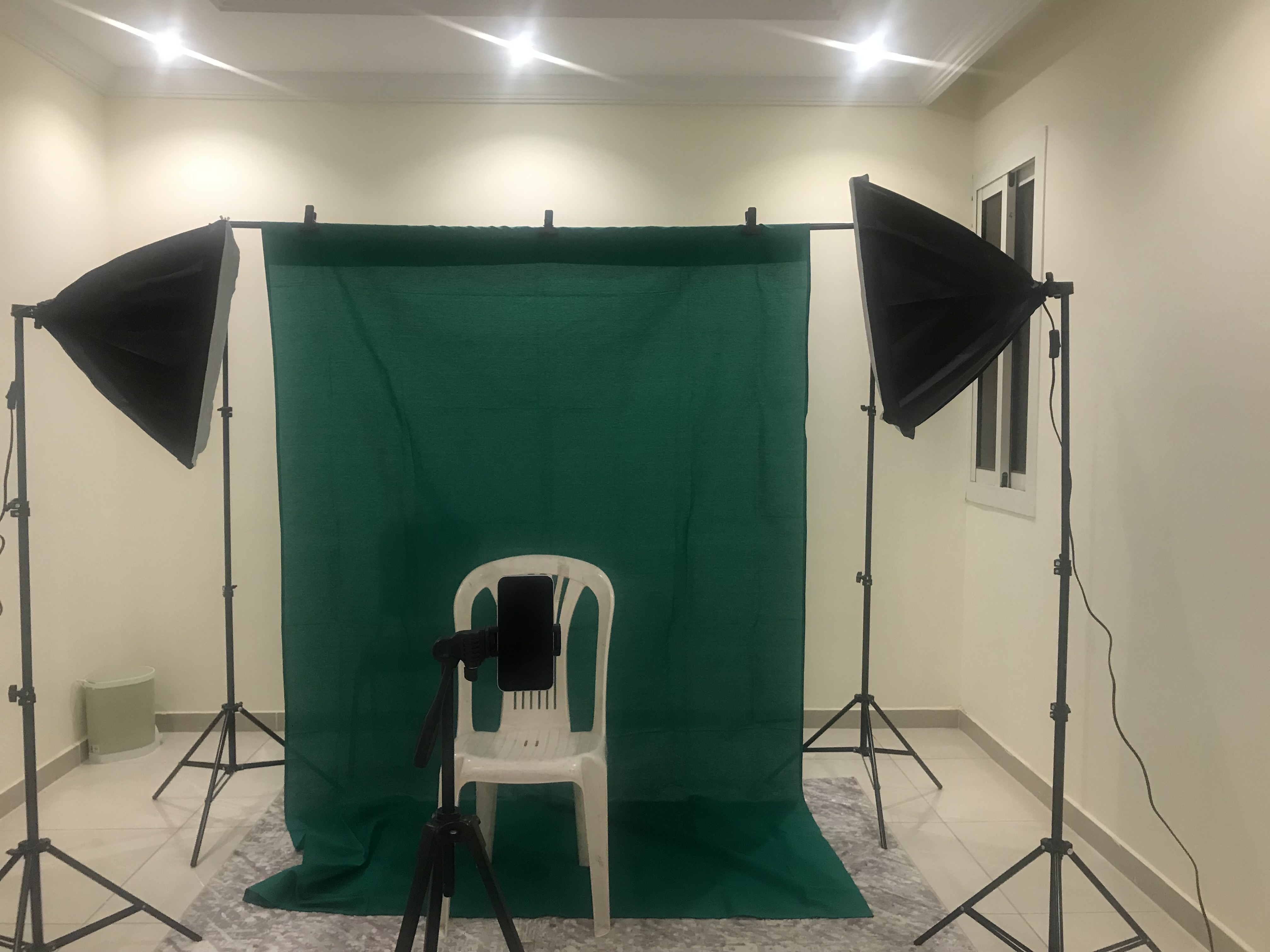}
    \caption{The setup of recording location}
    \label{fig:location}
\end{figure}

\paragraph{Phase 3: Record Videos}
The \textbf{recording setup} to guarantee the quality of the recordings consists of a controlled environment. A three meter tall and wide white and green fabric background was utilized to reduce distractions and increase the visibility of sign gestures. In addition, two softbox lights were strategically placed on both sides of the camera setup to ensure uniform and balanced illumination. The recording device was an iPhone 14 placed on a tripod approximately 2 meters distant from the signers. The signers were positioned at a distance of half meter from the fabric background. Figure~\ref{fig:location} illustrates the setup of the recording location.
\\
The recording process followed the method used in the How2Sign dataset~\cite{duarte2021how2sign}, as it provided a detailed and structured approach that closely aligned with the needs of our task. An expert translator first recorded the entire dataset, creating a video guide for participants to follow. Before recording each sentence, participants watched these videos to ensure they used the exact gestures and followed the same sequence. This consistency was crucial, as SL allows for multiple gestures or word orders to convey the same meaning. By using a standardized video guide, we ensured uniformity across all recordings. Additionally, to enhance the dataset's diversity and robustness, each signer repeated every sentence three times.
\paragraph{Phase 4: Process Data}
After recording, all videos were reviewed by sign language experts based on strict quality criteria, to ensure the quality of dataset. The criteria of accepting the recorded videos as follow: 
\begin{itemize}
\item The signs must be at regular speed.
\item Participants must perform the signs correctly.
\item The signs must be in the correct order based on the reference video.
\item Participants must not pause between words to perform the sentence.
\item All clips must be appropriate and free of laughter or talk
\item All clips must not contain any other signs than those specified in the reference video.
\end{itemize}
Accepted videos underwent post-processing, including resizing to 224x224 pixels, temporal normalization, and stabilization filters to enhance clarity.

\textbf{Temporal Dimension} The process of acquiring KAU-CSSL Dataset lasted approximately ten months. This time frame includes sentence compilation, expert consultation, participant recruitment, recording sessions, data verification, and final dataset preparation.
\textbf{Data Availability} The datasets gathered and analyzed during the current research is available upon request by the corresponding author. These data will be provided in a format that is appropriate for simplicity of use and accessibility, while also meeting with any relevant confidentiality or ethical standards.
\textbf{Dataset Naming Conventions} This dataset is titled "King Abdulaziz University - Continuous Saudi Sign Language (KAU-CSSL)" Dataset. This nomenclature encapsulates the dataset's purpose, regional context, and the nature of SL communication it contains.
\\

\paragraph{Dataset Scalability} The KAU-CSSL dataset has the ability to scale, ensuring that it can be used in real-world sign language recognition (SLR) applications. The dataset can be expanded to cover a larger range of vocabularies and diverse contexts by repeating the same systematic recording procedure used when building it. This involves recording multiple signers generating new signs and sentences and maintaining signing style, background, and context diversity. To accommodate domain-specific needs, such as medical, educational, or legal settings, the corpus may be supplemented with specially designed subsets in order to deal with vocabulary and situations of importance. The controlled recording setting and reference video ensure data quality consistency, and multiple signer inclusion and repeated recording ensure robustness and generalizability. This scalable approach allows the KAU-CSSL dataset to grow unproblematically, enabling the deployment of SLR systems that can be extremely tunable to different kinds of real-world applications. In conclusion, building and improving the dataset required meticulous planning, collaboration with subject matter experts, and extensive efforts to ensure the authenticity, relevance, and quality of the collected videos. The resulting dataset can facilitate meaningful communication between healthcare providers and the deaf community in medical contexts.

\section{Methodology}

The problem of continuous SSL recognition is formulated as a video classification task using a pre-trained VideoMAE model. Each video input is represented as a sequence of frames \( x = \{x_1, x_2, \ldots, x_{32}\} \), where each frame \( x_i \in \mathbb{R}^{224 \times 224 \times 3} \) corresponds to an RGB image of dimensions \( 224 \times 224 \). The model processes $32$ frames per video clip, which are evenly sampled from the video’s duration. Then, frames are normalized using mean and standard deviation values derived from ImageNet \cite{deng2009imagenet}. Data augmentations, including random horizontal flipping and rotation, are applied during training to enhance generalization. The target output \( y \) is a class label from a set of $85$ predefined classes, where each label represents a specific SL sentence (e.g., \textbf{oncologist}, \textbf{painful\_swallowing}). 
The KAU-CSSL dataset is split into 4,088 training, 870 validation, and 921 test videos. The model learns a function \( f \) that maps the input sequence to a probability distribution over the 85 classes, using a class-weighted cross-entropy loss to address minor class imbalances:

\[
\mathcal{L} = -\sum_{j=0}^{84} w_j y_j \log(p(y_j \mid x))
\]

where \( w_j \) is the weight for class \( j \), and \( p(y_j \mid x) \) is the predicted probability for class \( j \), obtained via a softmax function. The final prediction is the class with the highest probability:

\[
\hat{y} = \arg\max_{j \in \{0, \ldots, 84\}} \; p(y_j \mid x)
\]

\subsection{Dataset}
The used dataset is KAU-CSSL dataset, which is a dataset that was created for the recognition of Saudi Continuous Sign Language (SSL), it is composed of 5,879 RGB videos from 85 classes, consisting of medical and healthcare-related signs. The dataset is split into 4,088 training, 870 validation, and 921 test videos with an average sequence length of 118.8 frames. Videos were recorded at varying frame rates and resolutions and show various signers, including signers wearing niqabs, under varying lighting conditions and backgrounds. Classes such as \textbf{oncologist}, \textbf{painful\_slallowing}, and \textbf{cardiologist} represent domain-specific terminology. 

\subsection{Model Architecture}
\begin{figure}[htbp]
\centering
\scalebox{0.55}{
\begin{tikzpicture}[
    box/.style={rectangle, draw, rounded corners, minimum height=1cm, minimum width=1.5cm, align=center, font=\small},
    data/.style={rectangle, draw, minimum height=0.8cm, minimum width=2.0cm, align=center, fill=blue!10, font=\small},
    img/.style={rectangle, draw, minimum height=2cm, minimum width=4.0cm, align=center, font=\small},
    arrow/.style={-Stealth, thick},
    enc/.style={box, fill=green!20},
    lstm/.style={box, fill=orange!20},
    cls/.style={box, fill=purple!20},
    subbox/.style={rectangle, draw, rounded corners, minimum height=0.8cm, minimum width=1.2cm, align=center, font=\tiny},
    subdata/.style={rectangle, draw, minimum height=0.6cm, minimum width=1.5cm, align=center, fill=blue!5, font=\tiny},
    node distance=0.8cm and 1.2cm
]

\node[img] (frames) {\includegraphics[height=2cm]{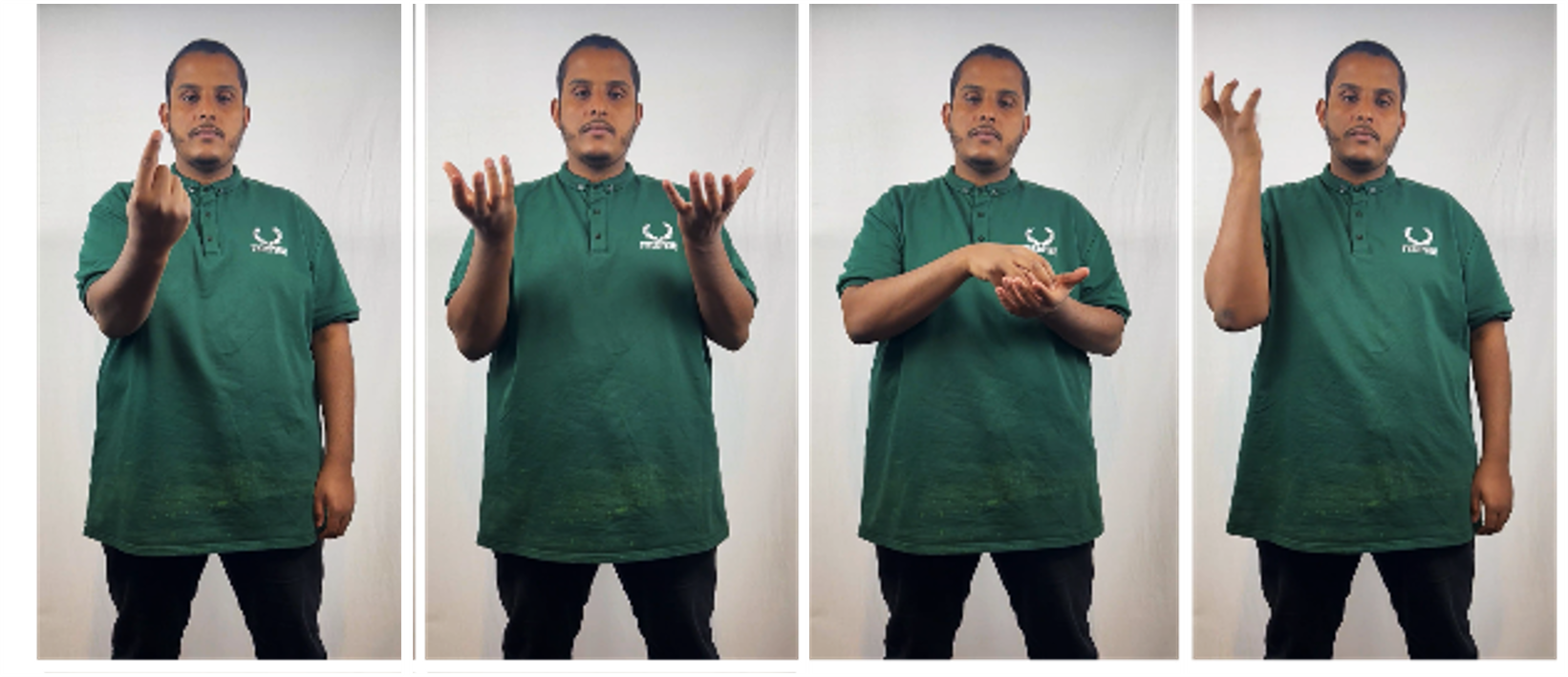}\\KAU-CSSL Frames};

\node[data, below=of frames] (input) {Video Input\\$T \times 224 \times 224 \times 3$\\($T = 16$ or $32$)};

\node[box, below=of input] (preproc) {Preprocessing\\(Sample, Resize, Normalize)};

\node[enc, below=of preproc] (resnet) {ResNet-18 Backbone\\(Pretrained on ImageNet)};
\node[data, below=of resnet] (features) {Frame Features\\$T \times 512$};

\node[box, below=of features] (proj) {Input Projection\\(Linear: 512 $\to$ 256)};
\node[data, below=of proj] (proj_out) {Projected Features\\$T \times 256$};

\node[box, below=of proj_out] (pos_enc) {Positional Encoding};
\node[data, below=of pos_enc] (pos_out) {Features with PE\\$T \times 256$};

\node[enc, below=of pos_out] (transformer) {Transformer Encoder\\3 Layers, 8 Heads\\(MHSA, FFN)};
\node[data, below=of transformer] (trans_out) {Transformed Features\\$T \times 256$};

\node[lstm, below=of trans_out] (lstm) {Bidirectional LSTM\\Hidden Size: 128};
\node[data, below=of lstm] (lstm_out) {LSTM Output\\$T \times 256$};

\node[cls, below=of lstm_out] (cls_head) {Classification Head\\(Mean Pool, Linear)};
\node[data, below=of cls_head] (output) {Class Probabilities\\85 Classes};
\node[right=0.2cm of output, font=\small] {$\mathcal{L}_{\text{CE}}$};

\draw[arrow] (frames) -- (input);
\draw[arrow] (input) -- (preproc);
\draw[arrow] (preproc) -- (resnet);
\draw[arrow] (resnet) -- (features);
\draw[arrow] (features) -- (proj);
\draw[arrow] (proj) -- (proj_out);
\draw[arrow] (proj_out) -- (pos_enc);
\draw[arrow] (pos_enc) -- (pos_out);
\draw[arrow] (pos_out) -- (transformer);
\draw[arrow] (transformer) -- (trans_out);
\draw[arrow] (trans_out) -- (lstm);
\draw[arrow] (lstm) -- (lstm_out);
\draw[arrow] (lstm_out) -- (cls_head);
\draw[arrow] (cls_head) -- (output);

\node[fit=(frames)(input)(preproc)(resnet)(features)(proj)(proj_out)(pos_enc)(pos_out)(transformer)(trans_out)(lstm)(lstm_out)(cls_head)(output), draw, thick, inner sep=0.5cm] (main_box) {};

\node[subdata, left=4.5cm of resnet] (resnet_sub_input) {Input\\$224 \times 224 \times 3$};
\node[subbox, fill=green!10, below=0.5cm of resnet_sub_input] (conv1) {Conv1\\7x7, 64, stride 2};
\node[subdata, below=0.5cm of conv1] (pool1) {After MaxPool\\$112 \times 112 \times 64$};
\node[subbox, fill=green!10, below=0.5cm of pool1] (res_block1) {Residual Block 1\\(3x3, 64 $\to$ 64)};
\node[subdata, below=0.5cm of res_block1] (res1_out) {Output\\$56 \times 56 \times 64$};
\node[subbox, fill=green!10, below=0.5cm of res1_out] (res_block2) {Residual Block 2\\(3x3, 64 $\to$ 128)};
\node[subdata, below=0.5cm of res_block2] (res2_out) {Output\\$28 \times 28 \times 128$};
\node[subbox, fill=green!10, below=0.5cm of res2_out] (res_block3) {Residual Block 3\\(3x3, 128 $\to$ 256)};
\node[subdata, below=0.5cm of res_block3] (res3_out) {Output\\$14 \times 14 \times 256$};
\node[subbox, fill=green!10, below=0.5cm of res3_out] (res_block4) {Residual Block 4\\(3x3, 256 $\to$ 512)};
\node[subdata, below=0.5cm of res_block4] (resnet_final_out) {Final Output\\$1 \times 1 \times 512$};
\draw[arrow] (resnet_sub_input) -- (conv1);
\draw[arrow] (conv1) -- (pool1);
\draw[arrow] (pool1) -- (res_block1);
\draw[arrow] (res_block1) -- (res1_out);
\draw[arrow] (res1_out) -- (res_block2);
\draw[arrow] (res_block2) -- (res2_out);
\draw[arrow] (res2_out) -- (res_block3);
\draw[arrow] (res_block3) -- (res3_out);
\draw[arrow] (res3_out) -- (res_block4);
\draw[arrow] (res_block4) -- (resnet_final_out);
\draw[arrow] (resnet.west) -- ++(-1.5cm,0) |- (resnet_sub_input.east);
\draw[arrow] (resnet_final_out.east) -- ++(1.5cm,0) |- (features.west);
\node[fit=(resnet_sub_input)(conv1)(pool1)(res_block1)(res1_out)(res_block2)(res2_out)(res_block3)(res3_out)(res_block4)(resnet_final_out), draw, dashed, inner sep=0.5cm, label={left:ResNet-18 Details}] (resnet_sub_box) {};

\node[subdata, right=4.5cm of transformer] (trans_sub_input) {Input\\$T \times 256$};
\node[subbox, fill=green!10, below=0.5cm of trans_sub_input] (mhsa) {MHSA\\8 Heads};
\node[right=0.2cm of mhsa, font=\tiny] {$+$};
\node[subbox, right=0.5cm of mhsa] (trans_norm1) {Layer Norm};
\node[subdata, below=0.5cm of mhsa] (mhsa_out) {Output\\$T \times 256$};
\node[subbox, fill=green!10, below=0.5cm of mhsa_out] (ffn) {FFN\\1024 Units};
\node[right=0.2cm of ffn, font=\tiny] {$+$};
\node[subbox, right=0.5cm of ffn] (trans_norm2) {Layer Norm};
\node[subdata, below=0.5cm of ffn] (trans_layer_out) {Output\\$T \times 256$};
\node[subdata, below=0.5cm of trans_layer_out] (trans_next_layer) {To Next Layer};
\draw[arrow] (trans_layer_out) -- (trans_next_layer);
\draw[arrow, dashed] (trans_next_layer.west) to[out=180,in=180,looseness=2] node[midway, left, font=\tiny] {$\times 3$ Layers} (trans_sub_input.west);
\node[subdata, below=0.5cm of trans_next_layer] (trans_final_out) {Final Output\\$T \times 256$};
\draw[arrow] (trans_next_layer) -- (trans_final_out);
\draw[arrow] (trans_sub_input) -- (mhsa);
\draw[arrow] (trans_sub_input) -- ++(0,-0.5cm) -| (trans_norm1.north);
\draw[arrow] (mhsa) -- (mhsa_out);
\draw[arrow] (mhsa_out) -- (ffn);
\draw[arrow] (mhsa_out) -- ++(0,-0.5cm) -| (trans_norm2.north);
\draw[arrow] (ffn) -- (trans_layer_out);
\draw[arrow] (transformer.east) -- ++(1.5cm,0) |- (trans_sub_input.west);
\draw[arrow] (trans_final_out.west) -- ++(-1.5cm,0) |- (trans_out.east);
\node[fit=(trans_sub_input)(mhsa)(trans_norm1)(mhsa_out)(ffn)(trans_norm2)(trans_layer_out)(trans_next_layer)(trans_final_out), draw, dashed, inner sep=0.5cm, label={right:Transformer Encoder Layer}] (trans_sub_box) {};

\node[subdata, left=4.5cm of lstm] (lstm_sub_input) {Input\\$T \times 256$};
\node[subbox, fill=orange!10, below=0.5cm of lstm_sub_input] (lstm_forward) {Forward LSTM\\Hidden Size: 128};
\node[subbox, fill=orange!10, below=0.5cm of lstm_forward] (lstm_backward) {Backward LSTM\\Hidden Size: 128};
\node[subbox, fill=orange!10, below=0.5cm of lstm_backward] (lstm_concat) {Concatenate Outputs};
\node[subdata, below=0.5cm of lstm_concat] (lstm_sub_output) {Output\\$T \times 256$};
\draw[arrow] (lstm_sub_input.south) -- (lstm_forward.north);
\draw[arrow] (lstm_forward.south) -- (lstm_backward.north);
\draw[arrow] (lstm_backward.south) -- (lstm_concat.north);
\draw[arrow] (lstm_concat.south) -- (lstm_sub_output.north);
\draw[arrow] (lstm.west) -- ++(-1.5cm,0) |- (lstm_sub_input.east);
\draw[arrow] (lstm_sub_output.east) -- ++(1.5cm,0) |- (lstm_out.west);
\node[fit=(lstm_sub_input)(lstm_forward)(lstm_backward)(lstm_concat)(lstm_sub_output), draw, dashed, inner sep=0.5cm, label={left:LSTM Details}] (lstm_sub_box) {};

\end{tikzpicture}
}
\caption{Model architecture of \texttt{KAU-SignTransformer} for SSL recognition}
\label{fig:Model_Architecture}
\end{figure}
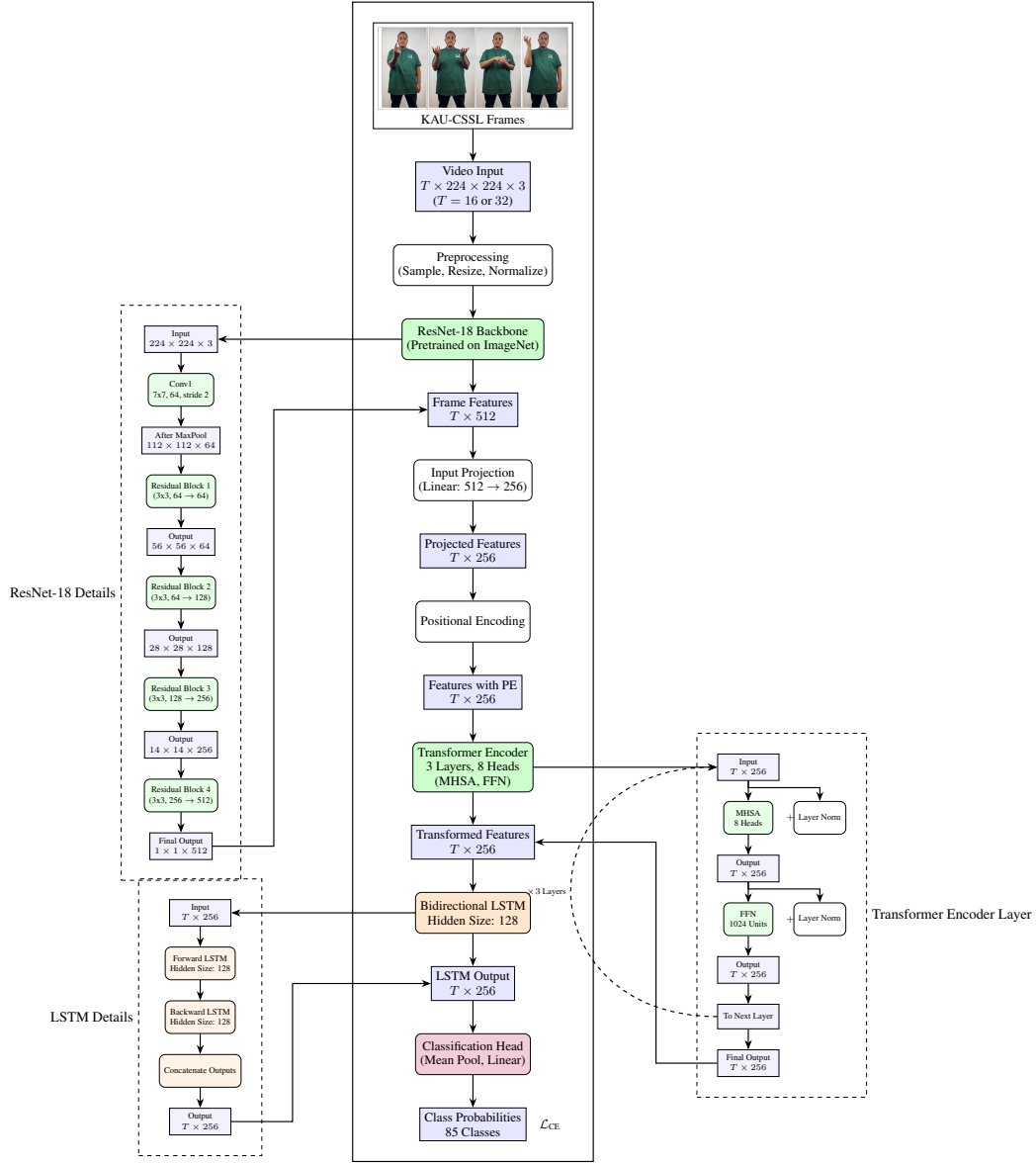

The \textbf{KAU-SignTransformer} model is designed for SSL video classification, with a series of modules integrated to process spatiotemporal features and predict gesture classes. The model consists of a ResNet-18 backbone, input projection layer, positional encoding, transformer encoder, bidirectional LSTM, and classification head. Ablation studies investigate variations: 16 vs. 32 frames, 0/1/3 transformer layers, unidirectional vs. bidirectional LSTM, presence/absence of positional encoding, pretrained vs. randomly initialized ResNet-18, and 8 vs. 16 attention heads. Figure ~\ref{fig:Model_Architecture} shows model architecture of \texttt{KAU-SignTransformer} for SSL recognition, with detailed views of ResNet-18, Positional Encoding, Transformer Encoder, and Bidirectional LSTM layers. The pipeline processes KAU-CSSL video frames vertically through a pretrained ResNet-18 backbone (featuring convolutional and residual blocks), input projection, positional encoding (adding sinusoidal encodings), transformer encoder (with MHSA, FFN, and layer normalization, stacked 3 times), bidirectional LSTM (with forward and backward layers), and a classification head to predict 85 SSL gestures.

\paragraph{Preprocessing}
Video frames are preprocessed by a series of preprocessing operations to prepare them for the model. In order to bring together the temporal axis, initially, \( T \) frames  32 are uniformly sampled from each video: linear interpolation over the video duration is performed with frame indices computed as \( \text{indices} = \text{linspace}(0, N-1, T) \), where \( N \) is the number of video frames. They are then resized to \( 224 \times 224 \) pixels using bilinear interpolation to match the input ResNet-18 backbone size.

Additionally, in training, data augmentations have been used for generalization enhancement: random horizontal flip with 0.5 probability, color jittering (brightness, contrast, and saturation each altered by as much as 10\%), and random rotation (as per ablation studies) introduce variability. All the frames are normalized into RGB format and normalized using ImageNet-derived statistics such that they align with pretrained ResNet-18 weights. The normalization rescales each pixel value \( \mathbf{p}_{c,h,w} \) in channel \( c \) (R, G, B) as:

\[
\mathbf{p}_{c,h,w}' = \frac{\mathbf{p}_{c,h,w} - \mu_c}{\sigma_c}
\]
where \( \mu = [0.485, 0.456, 0.406] \) and \( \sigma = [0.229, 0.224, 0.225] \) are the mean and standard deviation of R, G, and B channels, respectively. For testing and validation, resizing and normalization are performed without any augmentations to ensure consistent evaluation.

\paragraph{ResNet-18 Backbone}
The ResNet-18 backbone is a convolutional neural network (CNN) having 18 layers that is designed to mitigate the vanishing gradient problem using residual connections \cite{he2016deep}. It was chosen due to its compromise between depth and computationally efficiency with approximately 11.7 million parameters making it suitable for feature extraction for video tasks and pretrained weights on ImageNet \cite{deng2009imagenet}—a 1.28 million image dataset of 1000 classes—enabling good initialization for transfer learning by representing general visual features. ResNet-18 consists of an initial convolutional layer (7×7, 64 filters, stride 2), followed by a max-pooling layer (3×3, stride 2), reducing spatial resolution to 56×56. It then comprises four residual blocks with two convolutional layers (3×3 filters) with batch normalization and ReLU activation, and a shortcut connection summing the input to the output of the block, enabling the network to learn residual functions: \( \mathbf{y} = \mathbf{x} + \mathcal{F}(\mathbf{x}) \), where \( \mathcal{F} \) is the block transformation. Blocks progressively increase the numbers of filters (64, 128, 256, 512), and the shortcut in the first layer of every block (all except the first one) uses stride 2 in order to spatially downsample and generate feature maps of sizes 56×56, 28×28, 14×14, and 7×7, respectively. In the case when input and output sizes differ (e.g., when doubling filters), the shortcut includes a 1×1 convolution in order to match sizes. The global average pooling layer then reduces the final feature map of dimensions \( 512 \times 7 \times 7 \) to a vector of dimension 512 for every frame. Given as input a video tensor \( \mathbf{x} \in \mathbb{R}^{B \times T \times 3 \times 224 \times 224} \), where \( B \) is the batch size and \( T \) is the number of frames (16 or 32), the frames are reshaped into  \( B \cdot T \times 3 \times 224 \times 224 \) and fed into ResNet-18 without its final fully connected layer. Each frame is mapped into a 512-dimensional feature vector, getting a sequence of features \( \mathbf{F} \in \mathbb{R}^{B \times T \times 512} \), and then it's passed to the input projection layer for further calculation:
\\
\[
\mathbf{F}_{b,t} = \text{ResNet-18}(\mathbf{x}_{b,t}), \quad \text{for} \quad b \in \{1, \ldots, B\}, \quad t \in \{1, \ldots, T\}
\]

\( \mathbf{x}_{b,t} \in \mathbb{R}^{3 \times 224 \times 224} \) is the \( t \)-th frame of the \( b \)-th video, and \( \mathbf{F}_{b,t} \in \mathbb{R}^{512} \) is its feature representation.

\paragraph{Input Projection}
The ResNet-18 feature is projected to the dimension of the transformer by a linear layer. There exists a linear transformation for each feature vector \( \mathbf{F}_{b,t} \in \mathbb{R}^{512} \), that maps it into a \( d_{\text{model}} \)-dimensional space, where \( d_{\text{model}} = 256 \). This results in a sequence \( \mathbf{Z} \in \mathbb{R}^{B \times T \times 256} \). The projection is defined as:

\[
\mathbf{Z}_{b,t} = \mathbf{F}_{b,t} \mathbf{W}_{\text{proj}} + \mathbf{b}_{\text{proj}}
\]

where \( \\mathbf{W}_{\\text{proj}} \\in \\mathbb{R}^{512 \\times 256} \) and \( \\mathbf{b}_{\\text{proj}} \\in \\mathbb{R}^{256} \) are the weight matrix and bias of the linear layer, respectively.

\paragraph{Positional Encoding}
Positional encoding adds temporal order information to the projected features such that the transformer can determine the order of frames. The fixed sinusoidal encoding is employed, as in the standard transformer approach \cite{vaswani2017attention}. For each position \( t \) and dimension \( i \), the positional encoding \( \mathbf{PE} \in \mathbb{R}^{T \times 256} \) is computed as:
\[
\text{PE}(t, 2i) = \sin\left(\frac{t}{10000^{2i / d_{\text{model}}}}\right), \quad \text{PE}(t, 2i+1) = \cos\left(\frac{t}{10000^{2i / d_{\text{model}}}}\right)
\]

The encoded features are subsequently \( \mathbf{Z}' = \mathbf{Z} + \mathbf{PE} \), where \( \mathbf{Z}' \in \mathbb{R}^{B \times T \times 256} \), preserving the input dimensions but with temporal information embedded.

\paragraph{Transformer Encoder}
The transformer encoder projects the sequence \( \mathbf{Z}' \) to capture long-range temporal dependencies across frames. It contains 3 layers with 8 attention heads and a feed-forward network (FFN) of 1024 units. Every layer consists of multi-head self-attention (MHSA) followed by an FFN, with residual connections and layer normalization. In the case of a single layer, the MHSA operation for head \( h \) calculates attention scores as:
\[
\text{Attention}_h(\mathbf{Q}, \mathbf{K}, \mathbf{V}) = \text{softmax}\left(\frac{\mathbf{Q}_h \mathbf{K}_h^\top}{\sqrt{d_k}}\right) \mathbf{V}_h
\]

where \( \mathbf{Q}_h, \mathbf{K}_h, \mathbf{V}_h \in \mathbb{R}^{B \times T \times (d_{\text{model}}/H)} \) are projections for head \( h \), \( H = 8 \) is the number of heads, and \( d_k = d_{\text{model}}/H = 32 \). Outputs are concatenated and linearly projected, then followed by the FFN, resulting in \( \mathbf{T} \in \mathbb{R}^{B \times T \times 256} \).

\paragraph{Bidirectional LSTM}
A bidirectional Long Short-Term Memory (LSTM) layer also models temporal dependencies in the output of the transformer \( \mathbf{T} \). The LSTM has 128 hidden size per direction, resulting in a 256-dimensional output per timestep. In each direction (forward and backward), the LSTM updates its hidden state as:

\[
\mathbf{h}_t^{\text{fwd}} = \text{LSTM}_{\text{fwd}}(\mathbf{T}_t, \mathbf{h}_{t-1}^{\text{fwd}}), \quad \mathbf{h}_t^{\text{bwd}} = \text{LSTM}_{\text{bwd}}(\mathbf{T}_t, \mathbf{h}_{t+1}^{\text{bwd}})
\]

The forward and backward outputs are concatenated, resulting in \( \mathbf{L} \in \mathbb{R}^{B \times T \times 256} \), where \( \mathbf{L}_{b,t} = [\mathbf{h}_t^{\text{fwd}}; \mathbf{h}_t^{\text{bwd}}] \).

\paragraph{Classification Head}
The classification head pools the LSTM output \( \mathbf{L} \) by mean-pooling along the time axis, reducing it to a single vector per video, \( \mathbf{L}_{\text{mean}} \in \mathbb{R}^{B \times 256} \), where \( \mathbf{L}_{\text{mean},b} = \frac{1}{T} \sum_{t=1}^T \mathbf{L}_{b,t} \). A linear layer projects this to 85 class logits, with a softmax to obtain probabilities:

\[
\mathbf{p}_b = \text{softmax}(\mathbf{L}_{\text{mean},b} \mathbf{W}_{\text{fc}} + \mathbf{b}_{\text{fc}})
\]

where \( \mathbf{W}_{\text{fc}} \in \mathbb{R}^{256 \times 85} \) and \( \mathbf{b}_{\text{fc}} \in \mathbb{R}^{85} \) are the weight and bias of the final layer, and \( \mathbf{p}_b \in \mathbb{R}^{85} \) is the probability distribution over the 85 classes for the \( b \)-th video.
\subsection{Training Procedure}
The model was trained on the Aziz supercomputer using PyTorch. The objective function is a class-weighted cross-entropy loss. Training parameters are an 8 batch size, 50 epochs, and AdamW optimizer with learning rate 1e-4 and weight decay 1e-2. A cosine annealing learning rate scheduler scales the learning rate over epochs. Early stopping with patience of 10 epochs is applied on validation loss. Random horizontal flipping and rotation data augmentations enhance generalization. Configurations are trained independently in ablation studies, and random ResNet-18 initialization replaces ImageNet pretraining wherever needed.

\subsection{Evaluation Metrics}
Model performance was evaluated on KAU-CSSL test set , the perfromance was measured by overall accuracy and macro F1-score. Accuracy measures the proportion of correctly classified videos, whereas macro F1-score, the harmonic mean of precision and recall across classes averaged over classes, ensures balanced evaluation because of small class imbalances. Per-class accuracy, precision, recall, and F1-score are measured over a few representative classes to examine in-depth performance. Confusion matrices and classification reports are generated to examine misclassifications.

\section{Results}

Here, we present the result of our continuous SL recognition task with our transformer-based model. The model was trained on 85 classes. we evaluate the performance using different measures such as overall accuracy, precision, recall, F1-score, confusion matrix, and class-wise performance on test subsets of KAU-CSSL, which provides us with insights about strengths and weaknesses.

The data was split into three subsets, training, validation and testing, where the model was tested. The training subset was fully 4,085 samples which were used in the process of training. Validation subset is 827 samples which were selected for validation, enabling the selection of a model and hyperparameter early stopping thresholds. The test set of 940 videos across the 85 sign classes, it contains unseen samples which are accountable for model's generalization.

We are assessing the model based on the following metrics:
\begin{itemize}
    \item Accuracy: The proportion of correctly classified samples among total samples. \item Precision: The number of correctly predicted positive observations divided by the total predicted positives. \item Recall: The number of correctly predicted positive observations divided by all actual positives. \item F1-Score: The mean of precision and recall. \item Confusion Matrix: A matrix that displays the actual labels in comparison to the predicted labels, allowing us to identify frequent misclassifications.
\end{itemize}

\subsection{Model Performance}

\begin{figure}[]
\centering
    \includegraphics[width=15 cm]{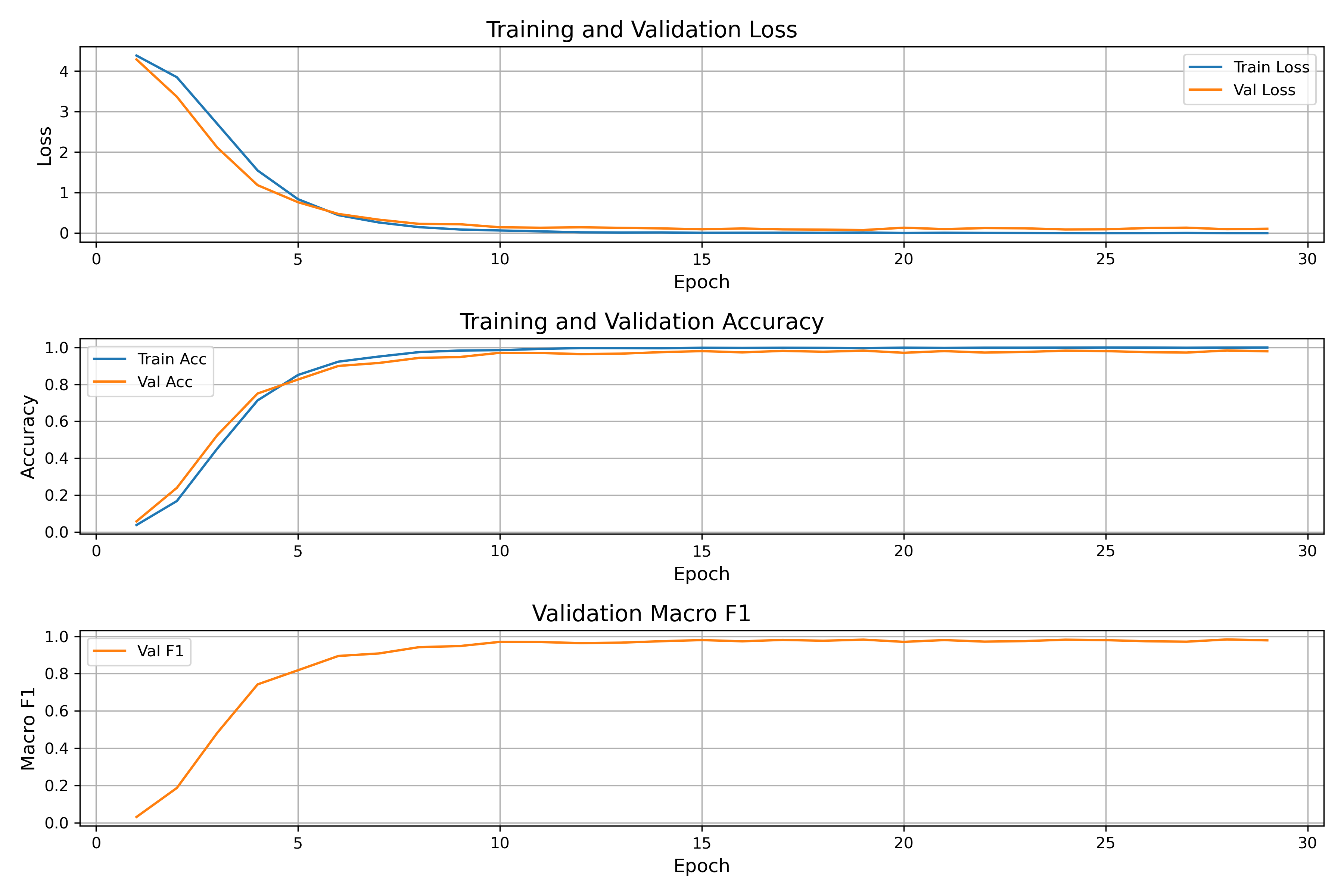}
    \caption{Training and Validation loss}
    \label{fig:results}
\end{figure}

A validation accuracy of 99.02\% in signer-dependent mode was achieved by the model in the testing phase. Early stopping began after 10 epochs with no improvements. High performance indicates ability to generalize well on the validation set. Figure~\ref{fig:results} shows training and validation loss over epochs.

The model attained the following overall performance values on the test set:
\begin{itemize}
    \item Test Accuracy: 99.02\%
    \item Macro-Average Precision: 99\%
    \item Macro-Average Recall: 99\%
    \item Macro-Average F1-Score: 99.01\%
\end{itemize}

As findings of this study indicate, the model generalizes well across a spectrum of SL classes, maintaining precision and recall aligned and extremely accurate.

Each of the 85 classes' performance metrics are tabulated in Table~\ref{tab:Classification_Report}. Observe how the model performed consistently well across the majority of classes, with precision and recall values between 90\% to 100\%. Some classes such as 'discount\_available' and 'medical\_report\_needed' had lower F1-scores due to their lower frequency in the dataset or the inherent difficulty in differentiation from the others.

The classification report table indicates that, the model performs extremely well in almost all classes except a few, achieving an F1-score of over 0.90 on 90\% of the classes. Classes like 'medical report needed' and 'discount available' are among those experiencing the drop in performance. This may be an indication of confusing these classes or some other similar hints in the dataset, thus showing the need for further improvement in dataset balancing or model fine-tuning.

\begin{figure}[]
\centering
    \includegraphics[width=0.97\linewidth]{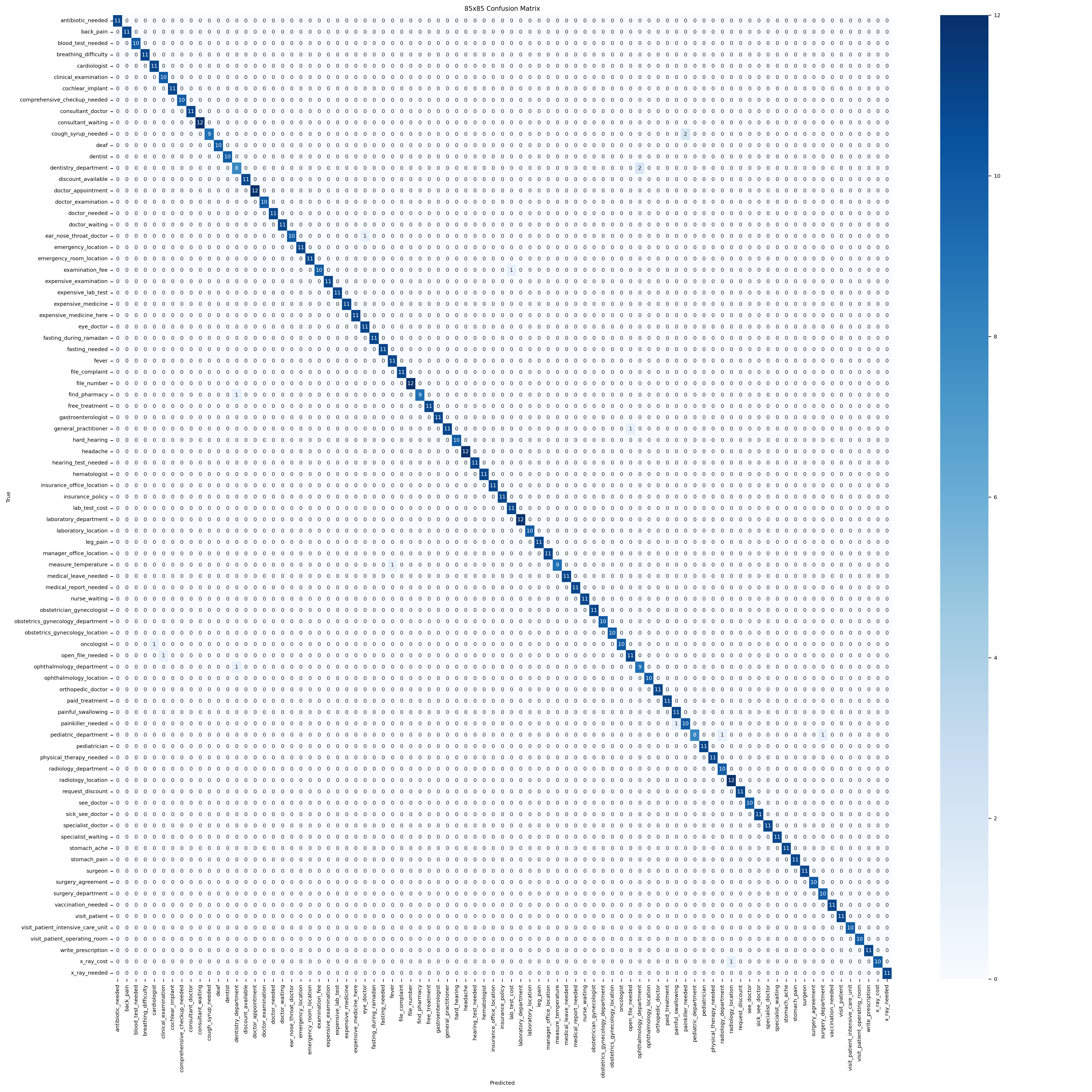}
    \caption{Confusion Matrix}
    \label{fig:confusion_matrix}
\end{figure}

\begin{figure}[]
\centering
    \includegraphics[width=9 cm]{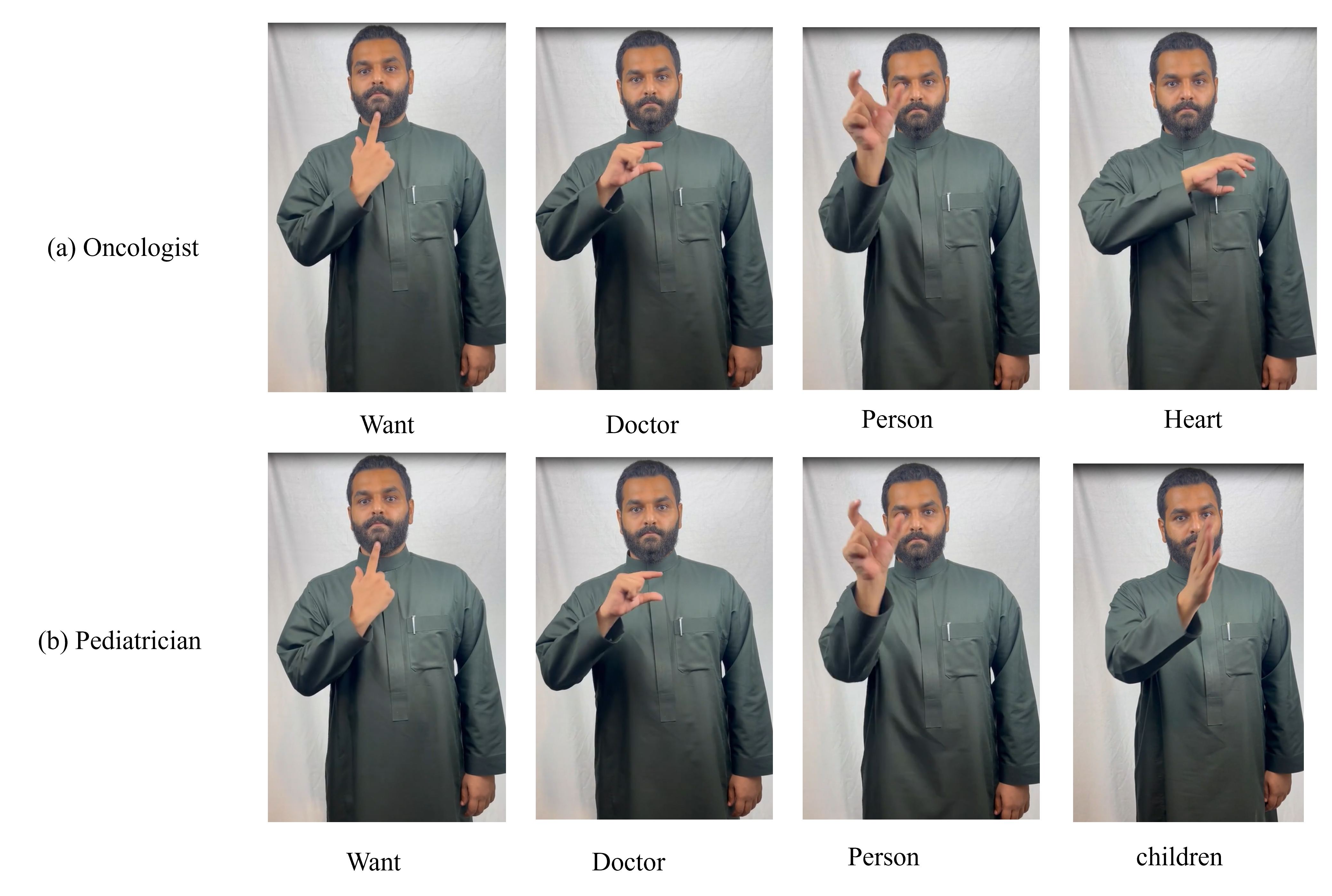}
    \caption{The similarity between 'Oncologist' and 'Pediatrician' sentences. (a) I want Oncologist (b) I want Pediatrician}
    \label{fig:compare}
\end{figure}

The confusion matrix gives even more information on the performance of classification on all 85 classes. It gives a diagonal entries that indicate correct classificatory results, and off-diagonal entries indicate misclassifications as can be seen in Fig. \ref{fig:confusion_matrix}.
There is high classification accuracy for most of the classes as marked by the prominent diagonal presence. However, there are certain confusions among classes having visually similar signs, such as 'Oncologist' and 'Pediatrician', or 'consultant waiting' and 'medical report needed' which may lead to misclassifications. These misclassifications may be due to common attributes of the sign gestures, which the model was unable to differentiate. Fig. \ref{fig:compare} shows the similarity of 'Oncologist' and 'Pediatrician'.

The performance of the model learning from the KAU-CSSL dataset is according to the following Table ~\ref{tab:performance_comparison}. In order to better detect complex hand movements and gestures in sign language, our model possessed a remarkable ability to build strong spatiotemporal representations from raw video input, as evidenced by its highest accuracy.
The competition results, demonstrating that it can capture gesture dynamics and movement well and effectively integrates spatial and temporal information efficiently. Below results demonstrate the KAU-CSSL dataset's generalizability and robustness and indicate that it is strong enough to support state-of-the-art transformer-based models.
\begin{table}[htbp]
\centering
\caption{Model Performance}
\label{tab:performance_comparison}
\begin{tabular}{lcccc}
\toprule
\textbf{Model} & \textbf{Accuracy (\%)} & \textbf{Precision (\%)} & \textbf{Recall (\%)} & \textbf{F1-score (\%)} \\
\midrule
KAU-SignTransformer       & 99.02 & 99 & 99 & 99.01 \\
\bottomrule
\end{tabular}
\end{table}

\begin{table}[htbp]
\centering
\caption{Classification Report} \label{tab:Classification_Report}
\resizebox{23em}{!}{
\begin{tabular}{|l|c|c|c|c|}
\hline
\textbf{Class} & \textbf{Accuracy} & \textbf{Precision} & \textbf{Recall} & \textbf{F1-score} \\ \hline
back\_pain & 1.0000 & 1.0000 & 1.0000 & 1.0000 \\ \hline
antibiotic\_needed & 1.0000 & 1.0000 & 1.0000 & 1.0000 \\ \hline
blood\_test\_needed & 1.0000 & 1.0000 & 1.0000 & 1.0000 \\ \hline
breathing\_difficulty & 1.0000 & 1.0000 & 1.0000 & 1.0000 \\ \hline
cardiologist & 1.0000 & 1.0000 & 1.0000 & 1.0000 \\ \hline
clinical\_examination & 1.0000 & 1.0000 & 1.0000 & 1.0000 \\ \hline
cochlear\_implant & 1.0000 & 1.0000 & 1.0000 & 1.0000 \\ \hline
comprehensive\_checkup\_needed & 1.0000 & 1.0000 & 1.0000 & 1.0000 \\ \hline
consultant\_doctor & 1.0000 & 1.0000 & 1.0000 & 1.0000 \\ \hline
consultant\_waiting & 1.0000 & 0.9231 & 1.0000 & 0.9600 \\ \hline
cough\_syrup\_needed & 1.0000 & 1.0000 & 1.0000 & 1.0000 \\ \hline
deaf & 1.0000 & 1.0000 & 1.0000 & 1.0000 \\ \hline
dentist & 1.0000 & 1.0000 & 1.0000 & 1.0000 \\ \hline
dentistry\_department & 0.9000 & 0.9000 & 0.9000 & 0.9000 \\ \hline
discount\_available & 1.0000 & 1.0000 & 1.0000 & 1.0000 \\ \hline
doctor\_appointment & 1.0000 & 1.0000 & 1.0000 & 1.0000 \\ \hline
doctor\_examination & 1.0000 & 1.0000 & 1.0000 & 1.0000 \\ \hline
doctor\_needed & 0.9091 & 1.0000 & 0.9091 & 0.9524 \\ \hline
doctor\_waiting & 1.0000 & 1.0000 & 1.0000 & 1.0000 \\ \hline
ear\_nose\_throat\_doctor & 1.0000 & 1.0000 & 1.0000 & 1.0000 \\ \hline
emergency\_location & 1.0000 & 1.0000 & 1.0000 & 1.0000 \\ \hline
emergency\_room\_location & 0.9091 & 1.0000 & 0.9091 & 0.9524 \\ \hline
examination\_fee & 1.0000 & 1.0000 & 1.0000 & 1.0000 \\ \hline
expensive\_examination & 1.0000 & 1.0000 & 1.0000 & 1.0000 \\ \hline
expensive\_lab\_test & 1.0000 & 1.0000 & 1.0000 & 1.0000 \\ \hline
expensive\_medicine & 1.0000 & 1.0000 & 1.0000 & 1.0000 \\ \hline
expensive\_medicine\_here & 1.0000 & 0.9167 & 1.0000 & 0.9565 \\ \hline
eye\_doctor & 1.0000 & 1.0000 & 1.0000 & 1.0000 \\ \hline
fasting\_during\_ramadan & 1.0000 & 1.0000 & 1.0000 & 1.0000 \\ \hline
fasting\_needed & 1.0000 & 1.0000 & 1.0000 & 1.0000 \\ \hline
fever & 1.0000 & 1.0000 & 1.0000 & 1.0000 \\ \hline
file\_complaint & 1.0000 & 1.0000 & 1.0000 & 1.0000 \\ \hline
file\_number & 1.0000 & 1.0000 & 1.0000 & 1.0000 \\ \hline
find\_pharmacy & 0.9000 & 1.0000 & 0.9000 & 0.9474 \\ \hline
free\_treatment & 1.0000 & 1.0000 & 1.0000 & 1.0000 \\ \hline
gastroenterologist & 1.0000 & 1.0000 & 1.0000 & 1.0000 \\ \hline
general\_practitioner & 1.0000 & 1.0000 & 1.0000 & 1.0000 \\ \hline
hard\_hearing & 1.0000 & 1.0000 & 1.0000 & 1.0000 \\ \hline
headache & 1.0000 & 1.0000 & 1.0000 & 1.0000 \\ \hline
hearing\_test\_needed & 1.0000 & 1.0000 & 1.0000 & 1.0000 \\ \hline
hematologist & 1.0000 & 1.0000 & 1.0000 & 1.0000 \\ \hline
insurance\_office\_location & 1.0000 & 0.9167 & 1.0000 & 0.9565 \\ \hline
insurance\_policy & 1.0000 & 1.0000 & 1.0000 & 1.0000 \\ \hline
laboratory\_department & 1.0000 & 1.0000 & 1.0000 & 1.0000 \\ \hline
laboratory\_location & 1.0000 & 1.0000 & 1.0000 & 1.0000 \\ \hline
lab\_test\_cost & 1.0000 & 1.0000 & 1.0000 & 1.0000 \\ \hline
leg\_pain & 1.0000 & 1.0000 & 1.0000 & 1.0000 \\ \hline
manager\_office\_location & 1.0000 & 1.0000 & 1.0000 & 1.0000 \\ \hline
measure\_temperature & 1.0000 & 1.0000 & 1.0000 & 1.0000 \\ \hline
medical\_leave\_needed & 1.0000 & 1.0000 & 1.0000 & 1.0000 \\ \hline
medical\_report\_needed & 0.9091 & 1.0000 & 0.9091 & 0.9524 \\ \hline
nurse\_waiting & 1.0000 & 1.0000 & 1.0000 & 1.0000 \\ \hline
obstetrician\_gynecologist & 1.0000 & 1.0000 & 1.0000 & 1.0000 \\ \hline
obstetrics\_gynecology\_department & 1.0000 & 1.0000 & 1.0000 & 1.0000 \\ \hline
obstetrics\_gynecology\_location & 1.0000 & 1.0000 & 1.0000 & 1.0000 \\ \hline
oncologist & 0.9091 & 0.9091 & 0.9091 & 0.9091 \\ \hline
open\_file\_needed & 1.0000 & 1.0000 & 1.0000 & 1.0000 \\ \hline
ophthalmology\_department & 0.9000 & 0.9000 & 0.9000 & 0.9000 \\ \hline
ophthalmology\_location & 1.0000 & 1.0000 & 1.0000 & 1.0000 \\ \hline
orthopedic\_doctor & 1.0000 & 1.0000 & 1.0000 & 1.0000 \\ \hline
paid\_treatment & 1.0000 & 1.0000 & 1.0000 & 1.0000 \\ \hline
painful\_swallowing & 1.0000 & 1.0000 & 1.0000 & 1.0000 \\ \hline
painkiller\_needed & 1.0000 & 1.0000 & 1.0000 & 1.0000 \\ \hline
pediatrician & 1.0000 & 0.9167 & 1.0000 & 0.9565 \\ \hline
pediatric\_department & 1.0000 & 1.0000 & 1.0000 & 1.0000 \\ \hline
physical\_therapy\_needed & 1.0000 & 1.0000 & 1.0000 & 1.0000 \\ \hline
radiology\_department & 0.9000 & 1.0000 & 0.9000 & 0.9474 \\ \hline
radiology\_location & 1.0000 & 1.0000 & 1.0000 & 1.0000 \\ \hline
request\_discount & 0.9091 & 0.9091 & 0.9091 & 0.9091 \\ \hline
see\_doctor & 1.0000 & 1.0000 & 1.0000 & 1.0000 \\ \hline
sick\_see\_doctor & 1.0000 & 1.0000 & 1.0000 & 1.0000 \\ \hline
specialist\_doctor & 1.0000 & 1.0000 & 1.0000 & 1.0000 \\ \hline
specialist\_waiting & 1.0000 & 1.0000 & 1.0000 & 1.0000 \\ \hline
stomach\_ache & 1.0000 & 1.0000 & 1.0000 & 1.0000 \\ \hline
stomach\_pain & 1.0000 & 1.0000 & 1.0000 & 1.0000 \\ \hline
surgeon & 1.0000 & 1.0000 & 1.0000 & 1.0000 \\ \hline
surgery\_agreement & 1.0000 & 1.0000 & 1.0000 & 1.0000 \\ \hline
surgery\_department & 1.0000 & 1.0000 & 1.0000 & 1.0000 \\ \hline
vaccination\_needed & 1.0000 & 0.9167 & 1.0000 & 0.9565 \\ \hline
visit\_patient & 1.0000 & 1.0000 & 1.0000 & 1.0000 \\ \hline
visit\_patient\_intensive\_care\_unit & 1.0000 & 1.0000 & 1.0000 & 1.0000 \\ \hline
visit\_patient\_operating\_room & 1.0000 & 1.0000 & 1.0000 & 1.0000 \\ \hline
write\_prescription & 1.0000 & 1.0000 & 1.0000 & 1.0000 \\ \hline
x\_ray\_cost & 1.0000 & 1.0000 & 1.0000 & 1.0000 \\ \hline
x\_ray\_needed & 1.0000 & 1.0000 & 1.0000 & 1.0000 \\ \hline
\textbf{Total} & \textbf{0.99} & \textbf{0.99} & \textbf{0.99} & \textbf{0.99} \\ \hline
\end{tabular}
}
\end{table}

The KAU-SignTransformer was evaluated in signer-independent mode, where the test set included unseen signers, achieving a test accuracy of 77.71\% compared to 99.02\% in signer-dependent mode as shown in Table ~\ref{tab:SignerDependent_SignerInependent}, the test set consists of unseen signers which made the task more complex, yet the model achieved good results.

\begin{table}[htbp]
\centering
\caption{Comparison of Signer Dependent \& Signer Independent results}
\label{tab:SignerDependent_SignerInependent}
\begin{tabular}{lcccc}
\toprule
\textbf{Mode} & \textbf{Accuracy (\%)} & \textbf{Precision (\%)} & \textbf{Recall (\%)} & \textbf{F1-score (\%)} \\
\midrule
Signer Dependent       & 99.02 & 99 & 99 & 99.01 \\
Signer Independent       & 77.71 & 83.47 & 79.86 & 78.30 \\
\bottomrule
\end{tabular}
\end{table}

\subsection{Ablation Study}
To assess the contribution of each element in our KAU-SignTransformer model to continuous sign language recognition, we conducted an ablation study on the KAU-CSSL dataset. We altered the essential architectural elements and training processes step by step and quantified their impact on performance in terms of test accuracy and macro-average F1-score. The factors under scrutiny are the pretrained ResNet-18 backbone, the Bidirectional LSTM for time handling, the Transformer Encoder layers, the number of attention heads, the data augmentations (random horizontal flip and color jittering), the number of input frames, and replacing ResNet-18 with ResNet-50. Table~\ref{tab:ablation_study} displays the outcome, providing us with an appreciation of the relative importance of each part.

\begin{table}[htbp]
\centering
\caption{Ablation Study Results for KAU-SignTransformer}
\label{tab:ablation_study}
\begin{tabular}{l S[table-format=2.2] S[table-format=2.2]}
\toprule
{Experiment} & {Accuracy (\%)} & {F1-Score (\%)} \\
\midrule
\textbf{Baseline}  & 99.02 & 99.01 \\
Reduce Number of Frames from 32 to 16 & 98.05 & 98.00 \\
Remove Transformer Encoder Layers (2 Layers) & 98.81 & 98.79 \\
Remove Transformer Encoder Layers (1 Layer) & 97.72 & 97.70 \\
Replace Bidirectional LSTM with Unidirectional LSTM & 98.81 & 98.79 \\
Using Randomly Initialized ResNet-18 Instead of Pretrained & 95.55 & 95.49 \\
Increase Transformer Attention Heads from 8 to 16 & 98.81 & 98.79 \\
Disable Data Augmentations & 98.70 & 98.67 \\
Replace ResNet-18 with ResNet-50 & 98.70 & 98.70 \\
\bottomrule
\end{tabular}
\end{table}

\paragraph{Effect of Reducing Number of Frames}
Reducing the input sequence size from 32 to 16 frames produced a test accuracy of 98.05\% and an F1-score of 98.00\%, a decrease of 0.97\% in accuracy compared to the baseline. This minor decrease suggests that the model can still learn relevant temporal information from fewer frames, though performance is somewhat reduced because there is less context available for gestures with longer transitions, such as "Movement Epenthesis" during continuous signing. \paragraph{Effect of Data Augmentations}
Data augmentations, including random horizontal-axis flipping and color jittering, were employed to encourage insensitivity to light and rotation changes. When these augmentations were removed, a test accuracy of 98.70\% and F1-score of 98.67\% was achieved, which was a 0.32\% decrease in accuracy from the baseline. This minimal loss indicates that augmentations confer a negligible generalization benefit, particularly for signs that have visual variability (e.g., "Antibiotic Needed"), but that the model remains very strong without them. \paragraph{Effect of Pretrained ResNet-18}
The baseline system uses a pretrained ResNet-18 for the learning of spatial features through transfer learning. With it replaced by a randomly initialized ResNet-18, there was a significant reduction in performance to 95.55\% accuracy and 95.49\% F1-score, representing a 3.47\% loss in accuracy. This suggests the significant contribution of pretrained weights in producing resilient feature representations that are vital to distinguish fine-grained SL gestures in the KAU-CSSL dataset. \paragraph{Effect of Substituting ResNet-18 with ResNet-50}
Substituting ResNet-18 for ResNet-50, a deeper network, yielded a test accuracy of 98.70\% and an F1-score of 98.70\%, a 0.32\% decrease in accuracy compared to the baseline. This shows that even though ResNet-50 gives more capacity, it does not significantly enhance performance on this task, perhaps due to the small size of the dataset or the adequacy of ResNet-18's feature extraction for SL recognition. \paragraph{Effect of LSTM for Temporal Modeling}
The Bidirectional LSTM captures temporal dependencies between frames, complementing the Transformer. Replacing it with a Unidirectional LSTM resulted in a test accuracy of 98.81\% and an F1-score of 98.79\%, a 0.21\% drop in accuracy. The minor reduction indicates that bidirectional processing is worth the effort, particularly for signs with complex sequences (e.g., "Radiology Department"), but the model remains excellent using unidirectional processing. \paragraph{Effect of Transformer Encoder}
Removing the Transformer Encoder altogether (relying on ResNet-18 alone and LSTM) is not tested but reducing layers does provide some understanding. Two layers had accuracy as 98.81\% and F1-score 98.79\% (0.21\% drop), while a single layer fell to 97.72\% and 97.70\% (1.30\% drop). This suggests the use of the self-attention mechanism in the Transformer is essential, with three layers achieving optimal performance for learning global frame interactions. \paragraph{Effect of Transformer Layers}
To test the impact of folding Transformer Encoder layers, we reduced it from three to one which lowered the accuracy to 97.72\% and F1-score to 97.70\%, a 1.30\% decrease. This indicates that an increased number of layers enhances the ability of the model to learn hierarchical representations, which is crucial in distinguishing subtle variations of signs for the 85-class. \paragraph{Effect of Increasing Transformer Attention Heads}
The effect of Increasing Transformer Attention Heads by double, where the attention heads goes from 8 to 16, made test accuracy 98.81\% and F1-score 98.79\%, resulting at a drop in accuracy of 0.21\%. This shows that having more heads does not necessarily improve the performance , possibly because there is redundancy or insufficient dataset complexity to make effective use of more heads. \paragraph{Summary of Ablation Results}
The ablation experiment indicates that the largest impact comes from pretrained ResNet-18, at 3.47\% loss of accuracy when randomly initialized, pointing to the importance of transfer learning at feature extracting phaze. We also observed that the transformer Encoder and layer count are also significant, the results shows that removing them cause up to 1.30\% drops, corresponding to their contribution in global context modeling. The Bidirectional LSTM and data augmentations yield small but noticeable improvements (0.21\% and 0.32\% decreases, respectively) in temporal and generalization capacities. Reducing frames to 16 and replacing ResNet-18 with ResNet-50 had minimal effects (0.97\% and 0.32\% decreases), pointing towards flexibility of such design decisions. These findings validate the KAU-SignTransformer architecture's resilience and direct potential improvements.
\section{Discussion}
The consistent performance across the training and test sets further supports the KAU-SignTransformer model's good generalizability to new data, as evidenced by its test accuracy of 99.02\% and F1-score of 99.01\% over the KAU-CSSL dataset.
This improved performance implies that the transfer learning approach, which makes use of a pretrained ResNet-18 backbone, successfully introduces strong temporal and spatial representations into the model, reducing overfitting and improving its capacity to handle a variety of sign patterns. This is further corroborated by the ablation study, which shows that transformer-based attention and pretrained features play a major role in achieving near-perfect recognition.

\subsection{Interpretation of Results}

The class-wise performance metrics provide important insights of the recognition capabilities on KAU-CSSL dataset at detailed level and clarify the influence of sign frequency and signers' behavior on performance results. We evaluated metrics of accuracy, precision, recall and F1-score and determined patterns related to characteristics of individual classes that clarifies the strengths and weaknesses of the model.

An important observation is the effect of common signs in all classes. Classes with high frequency, where signs appear in multiple sentences, such as Needed (appearing in multiple classes like "blood test needed" with 100\% accuracy and "Cough Syrup Needed" with 100\% accuracy), had higher accuracy and F1-scores. This is probably due to the fact that these common signs get more exposure during training, enabling the model to be trained more robustly on them. This robustness is supported by the ablation study’s finding of only a 0.97\% accuracy drop when reducing frames from 32 to 16, suggesting the model efficiently learns from frequent patterns.

Conversely, the model struggled to recognize rare classes, with signs appearing in one or two sentences (e.g., 'emergency room location' with 90.91\% accuracy ), which showed lower precision and recall. Although there is some difference between the classes results, class imbalance had no significant impact on the overall performance, since the extent of imbalance present in our dataset was quite slight, as evidenced by the near-perfect macro F1-score of 99.01\%. While rare signs are challenging due to limited representation, this high resilience indicates that the model can still cope with moderate variations in sign frequency.

The results were affected by the variability of gestures. Sentences considered "complex" where signs involving more than one gesture or performed using both hands,such as "Radiology Department" (90\% accuracy) and "Dentistry Department" (90\% accuracy), exhibited lower recognition accuracies,aligning with the ablation study’s 0.21\% drop when replacing Bidirectional LSTM with Unidirectional LSTM. These signs necessitate the model to understand complex movements and inter-hand relations which the contemporary methods of feature capture and temporal modeling is unable to completely solve. In contrast, “simple” sentences that use single-gestured signs like "Back Pain" and "Fever" (both 100\% accuracy) achieved perfect recognition, highlighting the model’s strength with less complex inputs.

Misclassification were rare but notable among classes with similar or overlapping gestures, 'how much' and 'department,' because shared hand shape or movement, For instance, "Emergency Room Location" (90.91\% accuracy) and "find pharmacy" (90\% accuracy) showed slight reductions, possibly due to overlapping hand shapes or movements, exacerbated by signer variability (e.g., "Consultant Waiting" with 92.31\% precision despite 100\% recall). The ablation study’s minimal 0.32\% drop when disabling data augmentations suggests the model is resilient to such variations, though there is style differences (e.g., speed, hand placement) that remain as a challenges.

The ablation study further proves these findings. The 3.47\% accuracy drop with a randomly initialized ResNet-18 underscores the pretrained backbone’s role in extracting robust spatial features, critical for distinguishing fine-grained gestures. Using one layer caused a 1.30\% drop which domnstrate the importance of transformer Encoder’s, indicating its capacity to capture global context, while the Bidirectional LSTM’s contribution is confirmed by the 0.21\% drop with a unidirectional replacement, emphasizing its role in sequential modeling.

These results suggest opportunities for refinement. The model’s high performance across most classes indicates a solid foundation, but addressing complex gestures and rare signs could involve targeted data augmentation or enhanced temporal modeling. The resilience to signer variability, supported by the ablation study’s findings on attention heads (0.21\% drop with 16 heads), suggests potential for further generalization with diverse training data.

These results bring to light several possibilities for the improvement of SL recognition systems. The low effects of class imbalance informs us that the model copes well with low variations in sign occurrence, although some scarce signs might need additional attention, such as through the use of data augmentation by synthetic gesture creation. Increases in misclassification due to higher order motions performed along with some basic sign suggest the development of more precise feature extraction for hand configurations and movements. In addition, the variability of the signers implies that many differences in style could be helpful at training datasets to build sophisticated model and customization techniques.

The performance metrics at the class were composed indicate that the success of recognition is defined by the signs frequency, the gesture's complexity, and variability of the signer. Common signs have the advantage of being repeated across signs, whereas complicated signs, infrequent signs, overlapping movements still remain problems. They open possibilities for future improvements by changing feature representation and by incorporating contextual information to design more precise and less restrictive sign language recognition systems. Resolving these problems will enable smoother communication for the Deaf and hard-hearing community.

The 77.71\% accuracy in signer-independent mode, compared to 99.02\% in signer-dependent mode, reflects challenges in generalizing to new signers, likely due to variations in signing style.
The results of this study yield a sign that the proposed method is very effective in detecting SL using continuous video data. Yet, there is more room for improvement such as enhancing the model to be more robust and accurate.

\subsection{Challenges and Limitations}

Several challenges were encountered during the dataset’s development, and during model development influencing the performance: \textbf{Sign Data Obscured by Niqab}
The niqab: Is the face garment that covers most of the face. Sign data Obscured by Niqab illustrates difficulties in capturing all useful features, specifically those involved the face where important facial expressions gets obscured. This challenge requires careful camera positioning and lighting to balance data quality and participant comfort. \textbf{Exclusion of Colored Gloves} Unlike other SL datasets using colored gloves for gesture visibility, the KAU-CSSL relied on natural hand movements, necessitating precise hand orientation and motion capture. The other datasets usually include the use of colored gloves which makes the gestures more conspicuous but not practicable for real life applications. This dataset focus on real life setting witch require the need for more robust models that accurately recognize the detail in the movement as well as the orientation of the hands. \textbf{Lack of Depth Data}  The dataset lacks depth information, which considered an important information to capture the three-dimensional aspect of the sign. \textbf{Feature extraction from a complex background} Green and white backgrounds, without colored gloves, increased the difficulty of feature extraction compared to uniform backgrounds, requiring robust segmentation. \textbf{Large vocabulary and Scalability}
With 85 classes, the dataset’s vocabulary poses misclassification risks, a challenge that scales with larger SL dictionaries. \textbf{Ambiguous Signs}
There were minimal misclassifications, almost wholly between very accurate differences in signs between classes, even though the model was good enough in most instances for all classes we still might cope with some issues of unbalanced and vague predictions for a class. \textbf{Movement Epenthesis (ME)} Movement Epenthesis (ME) is a term for the transition signs that occur whenever the hand moves between signs, it happens when the hand moves from the end of one sign to the beginning of another. This is possible frequent when signing sentences, sometimes such transitions are as lengthy as the signs themselves such as in "Radiology Department" (90\% accuracy), where the model has to differentiate between large gestures and transitional movements, as indicated by the 0.97\% frame reduction drop. \textbf{Signer-Dependent variations} People sign a bit differently. Variation in signature style (e.g., speed or hand position) affected classes like "Consultator Waiting" (92.31\% accuracy), which was hard to generalize even with the overall robustness of the model. \textbf{Occlusion} SL recognition models rely on a signer's hand and face characteristics. The occlusion problem, however, arises in SL datasets when the signer's hands occlude each other or the signer's face, whose performance can deteriorate on "Dentistry Department" (90\% accuracy), hindering feature extraction. \textbf{Sub sign reasoning} Some of the signs share similar movement or patterns, and there are supersigns made up of small "sub-signs" inside them. This complicates continuous SL recognition as the system has to determine how to handle these nested signs in an appropriate manner and necessitate accurate boundary definition, resulting in challenging continuous recognition.  \textbf{Short sign detection} Signs does not have equal length, where some are noticeably shorter than others. Which means fewer frames of data for those short signs like "doctor needed" (90.91\% accuracy), which can lead to risk of data loss, making their recognition more challenging, as reflected in the frame reduction analysis. Signs are not equally long and thus have varying frame counts, with signs being much shorter than others. That translates to lower frame counts of data for shorter signs like "doctor needed" (90.91\% accurate), which is a cause of data risk of loss, with their detection becoming increasingly challenging, as shown in the frame reduction analysis.

\subsection{Future Work}

Future work will focus on incorporating the domain-specific vocabularies into the KAU-CSSL dataset. Signer-independence tests will be conducted to analyze the method's cross-environment generalizability to support real-world deployment across multiple applications.
Future work will evaluate our model's performance with previous layers trained on available sign language datasets such as ASL, BSL, or CSL, and fine-tuning the last classifier on the KAU-CSSL dataset. The approach can be applied to improve model generalization and recognition performance by exploiting the common features between different sign languages.
\section{Conclusions}
This paper introduces the KAU-CSSL dataset and the KAU-SignTransformer model, advancing continuous Saudi Sign Language (SSL) recognition.
This dataset is an important progression in the domain of SL recognition, specifically for the SSL community due the absence of resources related to continuous SL recognition in the Saudi context. The dataset is based on medical sentences to improve the quality and accessibility of healthcare services for the deaf and hard of hearing community in Saudi Arabia through the examination of medical communication scenarios, which is our primary objective.
The dataset was collected with the incorporation of a wide variety of signers, including individuals with varying skin tones, genders, and niqab-wearing women. The dataset was focused on representing the diversity of the SSL community. 

The KAU-SignTransformer, leveraging transfer learning, achieved an outstanding 99.02\% accuracy and 99.01\% F1-score, validating its efficacy. The ablation study confirmed the pretrained ResNet-18, Transformer Encoder, and Bidirectional LSTM as key contributors, guiding future optimizations. Our research highlights the importance of tailored SL resources, promoting equitable communication. Future work will refine the model and dataset, addressing challenges like complex gestures and signer variability, to further empower accessibility technologies.
Additionally, the model’s 77.71\% accuracy in signer-independent mode underscores its potential for real-world scenarios, though further enhancements are needed to address signer variability.

In conclusion, our research emphasizes the significance of focusing on the development of resources that specifically address the requirements of individuals who are deaf or hard of hearing. By increasing the availability of SL resources, specifically for SSL, we make progress toward reaching equal communication opportunities for all individuals, irrespective of their hearing capabilities.

\section*{Acknowledgments}
The computation for the work presented in this paper was supported by KAU's High Performance Computing Center (Aziz Supercomputer).

\bibliographystyle{unsrtnat}

\end{document}